%% file: main.tex
\documentclass[sigconf]{acmart}
\AtBeginDocument{%
  \providecommand\BibTeX{{%
    \normalfont B\kern-0.5em{\scshape i\kern-0.25em b}\kern-0.8em\TeX}}}

\copyrightyear{2024}
\acmYear{2024}
\setcopyright{rightsretained}
\acmConference[HT '24]{35th ACM Conference on Hypertext and Social Media}{September 10--13, 2024}{Poznan, Poland}
\acmBooktitle{35th ACM Conference on Hypertext and Social Media (HT '24), September 10--13, 2024, Poznan, Poland}
\acmDOI{10.1145/3648188.3675141}
\acmISBN{979-8-4007-0595-3/24/09}

\usepackage{framed}
\usepackage{tabularx}
\usepackage{graphicx}
\usepackage{multirow}
\usepackage{color, colortbl}
\usepackage{subfigure}

\newcommand{\control}{\textbf{\texttt{Control}}}
\newcommand{\tutorialrandom}{\textbf{\texttt{Debugging-R}}}
\newcommand{\tutorialdec}{\textbf{\texttt{Debugging-D}}}
\newcommand{\tutorialinc}{\textbf{\texttt{Debugging-I}}}

\newcommand{\ignore}[1]{}
\newcommand{\tabincell}[2]{\begin{tabular}{@{}#1@{}}#2\end{tabular}}
\usepackage{xspace}
\newcommand{\etal}{\emph{et al.}\xspace}
\newcommand{\paratitle}[1]{\vspace{1.0ex}\noindent\textbf{#1}}
\newcommand{\ie}{\textit{i.e.,}~}
\newcommand{\eg}{\textit{e.g.,}~}

\begin{document}

\title{To Err Is AI! Debugging as an Intervention to Facilitate Appropriate Reliance on AI Systems}
\author{Gaole He}
\affiliation{%
 \institution{Delft University of Technology}
 \city{Delft}
 \country{The Netherlands}
 }
\email{g.he@tudelft.nl}

\author{Abri Bharos}
\affiliation{%
 \institution{Delft University of Technology}
 \city{Delft}
 \country{The Netherlands}
 }
\email{a.r.j.bharos@gmail.com}

\author{Ujwal Gadiraju}
\affiliation{%
 \institution{Delft University of Technology}
 \city{Delft}
 \country{The Netherlands}
 }
\email{u.k.gadiraju@tudelft.nl}

\renewcommand{\shortauthors}{Gaole He, Abri Bharos, and Ujwal Gadiraju}


% \settopmatter{printacmref=false}

%Abstract EDITED

\begin{abstract}

Powerful predictive AI systems have demonstrated great potential in augmenting human decision making. Recent empirical work has argued that the vision for optimal human-AI collaboration requires `\textit{appropriate reliance}' of humans on AI systems. However, accurately estimating the trustworthiness of AI advice at the instance level is quite challenging, especially in the absence of performance feedback pertaining to the AI system. 
{In practice, the performance disparity of machine learning models on out-of-distribution data makes the dataset-specific performance feedback unreliable in human-AI collaboration.}
% In practice, the out-of-distribution data is a common issue where machine learning models may suffer from performance disparity and the performance feedback from pre-defined test set will no longer be reliable. 
% the performance feedback from pre-defined dataset may fail to faithfully reflect the reliability of AI system.
Inspired by existing literature on critical thinking and a critical mindset, we propose the use of debugging an AI system as an intervention to foster appropriate reliance. 
In this paper, we explore whether a critical evaluation of AI performance within a debugging setting can better calibrate users' assessment of an AI system %~\glcomment{user estimation / assessment of AI performance?}
and lead to more appropriate reliance. 
Through a quantitative empirical study ($N=234$), we found that our proposed debugging intervention does not work as expected in facilitating appropriate reliance. Instead, we observe a decrease in reliance on the AI system after the intervention --- potentially resulting from an early exposure to the AI system's weakness. %it may suffer from a bad first impression brought by early exposure of AI system's weakness. 
% can better calibrate user perceptions of AI performance. 
We explore the dynamics of user confidence {and user estimation of AI trustworthiness across groups with different performance levels} to help explain how inappropriate reliance patterns occur. 
Our findings have important implications for designing effective interventions to facilitate appropriate reliance and better human-AI collaboration. %\ujcomment{update abstract based on findings!}
\textit{\textbf{This is an expanded version of HT'24 paper, providing more details and experimental analysis.}}
\end{abstract}

\maketitle

\input{sections/sec-intro}
\input{sections/sec-rel-new}
% \input{sections/sec-rel}
\input{sections/sec-task}

\input{sections/sec-exp}
\input{sections/sec-result}
\input{sections/sec-discussion}
\input{sections/sec-con}

% \begin{acks}
% \end{acks}

\bibliographystyle{ACM-Reference-Format}
\bibliography{critical}

\input{sections/appendix}

\end{document}

%% file: sections/sec-intro.tex
\section{Introduction}
% \textcolor{blue}{Background human-AI collaboration, miscalibrated trust, inappropriate reliance}
% \glcomment{To improve, current story is still not very sound}
With the rise of deep learning systems over the last decade, there has been a widespread adoption of AI systems in supporting human decision makers~\cite{lai2021towards}, albeit without always fully understanding the societal impact or downstream consequences of relying on such systems~\cite{erlei2020impact,erlei2022s}. 
Due to the opaqueness of some AI systems, users have struggled to determine when exactly they are trustworthy and have failed to achieve a complementary team performance. As a result, several previous studies that have explored human-AI collaboration and teaming across different contexts have reported improvements over human performance stemming from AI assistance, although this often falls short of AI performance~\cite{bansal2021does,Liu-CSCW-2021}.
To realize the full potential of complementary team performance, human decision makers need to identify when they should rely on AI systems (\ie identifying instances where AI systems are capable or more capable than humans) and when they are better off relying on themselves (\ie identifying instances where AI systems are less capable than humans). Such a reliance pattern has been defined as \textit{appropriate reliance}~\cite{Lu-CHI-2021,lai2021towards}. %In a series of recent study~\cite{Lu-CHI-2021,lai2021towards}, such ideal reliance patterns is called appropriate reliance.

In practice, it is common that users need to deal with data from unknown distributions and unseen contexts, meaning that AI systems in the real-world need to provide users with advice on out-of-distribution data~\cite{Liu-CSCW-2021,chiang2021you}. 
Under such circumstances, the estimated performance of an AI system or the so-called `stated accuracy' of the system (\ie accuracy on pre-defined test sets) cannot faithfully reflect the trustworthiness of the AI system. 
% Meanwhile accuracy and related measures are also not ideal indicator ~\cite{papenmeier2022accurate}, 
Only a few works~\cite{Lu-CHI-2021} have explored how humans rely on AI systems when performance feedback is limited or scarce. Previous work has found that user agreement with AI advice in tasks where they have high confidence significantly affects their reliance on the system, in the absence of the stated accuracy or performance of the system~\cite{Lu-CHI-2021}. 
% ``agreement between people and a model on decision-making tasks that people have high confidence in signicantly affects reliance on the model if people receive no information about the model’s performance''. 
To help users assess the trustworthiness of AI systems, a practical solution that has been proposed, is to provide meaningful explanations along with AI advice~\cite{LIME-KDD-2016,toreini2020relationship}. Post-hoc explanations have been found to improve user understanding of AI advice in empirical studies exploring human-AI decision making~\cite{wang2021explanations,lai2021towards}. 
% explanations alone
However, most existing XAI methods have remained ineffective in helping users assess the trustworthiness of AI advice at the instance level, adversely affecting the degree of appropriate reliance of users on AI systems~\cite{chandrasekaran2018explanations,wang2021explanations}. %and further promote appropriate reliance on AI systems.~\glcomment{To check whether we should put critical mindset as motivation.}

To realize the goal of appropriate reliance, human decision makers need to be capable of evaluating AI advice and the trustworthiness of the AI system critically. We argue that such a critical mindset can help users avoid blindly following AI advice (\ie avoiding \textit{over-reliance}), and also prevent them from distrusting AI advice when it can be productive (\ie avoiding \textit{under-reliance}).
Inspired by recent works on explanation-based human debugging of AI systems~\cite{lertvittayakumjorn2021explanation,balayn2022can}, we propose explanation-based debugging as a training intervention to increase appropriate reliance on AI systems. 
% we propose to involve users in a user tutorial composed of a video and explanation-based debugging phase. 
We posit that such a debugging intervention has the potential to help users understand the limitations of AI systems --- \textbf{\textit{that neither explanations of the AI advice nor the advice itself are always reliable}}. 
Recognizing these limitations %and combining with domain-specific knowledge (\eg task guidelines)~\glcomment{shall we mention task guidelines here? Or it's just one design to make users more competitive in the task}, 
can help users better understand when an AI system is trustworthy and thereby increase appropriate reliance on the system. In this paper, we aim to empirically evaluate the effectiveness of using a debugging intervention as a means to increase appropriate reliance and address the following research questions.
% To develop critical evaluation skills on specific decision making tasks, we propose to involve users in debugging AI advice based on explanations. 
% \textcolor{blue}{Describe RQs}
%To analyze the impact of perceived trustworthiness of AI systems on user reliance when performance feedback is limited, we aim to find answers for the following two research questions:~\glcomment{Shall we include ``when performance feedback is limited'' in RQ?}
\begin{framed}
%\textbf{RQ1:} How does the perceived trustworthiness of AI systems shape user reliance on AI systems when performance feedback is limited?
%\noindent\textbf{RQ1:} How can the confidence of a user in an AI system at the instance level explain user reliance on the system? \\
\noindent\textbf{RQ1:} How can a debugging intervention help users to estimate the performance of an AI system, both at the instance and at the global level? \\
%\textbf{RQ2:} Can we promote appropriate reliance  with a critical evaluation of trustworthiness of AI systems?
\textbf{RQ2:} How does a debugging intervention affect the reliance of users on an AI system?
% How can the Dunning-Kruger Effect be mitigated in human-AI decision making tasks?
\end{framed}
% \glcomment{The frame seems to occupy some space, we can use pure text to save space when necessary.}
% \textcolor{blue}{describe our methods and results} 
To answer the above questions, we propose three hypotheses considering the effect of the debugging intervention on AI performance assessment as well as reliance, and the task ordering effect of debugging intervention on appropriate reliance. 
% \ujcomment{use `debugging intervention' consistently; no need to use training or tutorial} on appropriate reliance. 
We tested these hypotheses in an empirical study ($N$ = 234) of human-AI collaborative decision making in a deceptive review detection task (\ie identifying whether one piece of review is written based on real experience). Interestingly, however, we found that the proposed debugging intervention fails to calibrate user estimation of AI performance and further promote appropriate reliance. 
% We found that the proposed debugging intervention is effective in improving both AI performance assessment and appropriate reliance on AI systems. 

% \textcolor{blue}{highlight key findings and implications (contribution)}
Our results highlight that when presented with the weakness of the AI system in an early stage of the debugging intervention, users underestimate AI performance and rely less on the AI system. 
Users' overestimation of their own competence may further amplify such an effect. 
%Our findings provide useful implications to mitigate such cognitive bias. Meanwhile, 
We analyzed user confidence evolution across the different reliance patterns exhibited, which helps explain why inappropriate reliance occurs. {Through an analysis exploring relatively less-competent individuals, we found that the underestimation of AI trustworthiness may also play a role in shaping under-reliance, which is potentially relevant to the metacognitive bias called the Dunning-Kruger effect~\cite{kruger1999unskilled}.} 
Our work has important implications for designing effective interventions to promote appropriate reliance in the context of human-AI decision making.
% \ujcomment{Our work has important implications on ...}~\glcomment{One sent is added before the comment}

%% file: sections/sec-rel-new.tex
\section{Related Work}
Our work is closely related to the studies on human-AI decision making, %~\glcomment{Do you think AI-assisted decision making would be more clear?},
appropriate reliance on AI systems, and explanation-based debugging of machine learning systems.
% ~\glcomment{To save space, we can finally modify the subsections into paratitles}

\paratitle{Human-AI Decision making}. 
% \begin{itemize}
%     \item With the rise of ML, AI systems have been widely used. However, due to ethical and multiple consideration, full automation is undesirable. Therefore, AI system played a role of assistant in supporting human decision makers. 
%     % In this context, researchers have been putting more efforts for human-AI decision making. 
%     \item Existing work mainly explored some factors will affect user trust and reliance on AI systems
% \end{itemize}
With the technical advances of deep learning methods in the recent decade, researchers have shown much interest in putting such methods for a wide arrange of applications (like medical image analysis~\cite{litjens2017survey}, autonomous driving~\cite{grigorescu2020survey}). 
However, due to the intrinsic uncertainty and opaqueness of such AI systems, it would be undesirable to make AI systems automate the decision making, especially in high-stakes scenarios (\eg legal judgment, medical diagnosis). 
Under such circumstances, AI systems are expected to play a supporting role for human decision makers. 
According to GDPR, users have the right to obtain meaningful explanations to work with such AI systems~\cite{selbst2018meaningful}. 
Motivated by this, a series of work has proposed to construct human-centered explainable AI systems~\cite{ehsan2020human,ehsan2021operationalizing,liao2021human} for better human-AI collaboration. 
% % and conducted empirical work about how explanations will shape user trust and reliance~\cite{wang2021explanations}. 
%~\glcomment{Do you think the human-centered XAI should not be mentioned here?}
Existing work has widely explored how different user factors (\eg expertise~\cite{Nourani-HCOMP-2020,dikmen2022effects}, risk perception~\cite{green2020algorithmic}, machine learning literacy~\cite{Chiang-IUI-2022}) and interaction designs (\eg performance feedback~\cite{bansal2019beyond,yin2019understanding,Rechkemmer-CHI-2022}, explanation~\cite{wang2021explanations}, user tutorial~\cite{Lai-CHI-2020,mozannar2022teaching}) will affect user trust in and reliance on AI systems.

\paratitle{Appropriate Reliance on AI Systems}. One important goal of human-AI decision making is complementary team performance~\cite{Liu-CSCW-2021,bansal2021does}, which requires appropriate reliance~\cite{lee2004trust}. 
In practice, however, humans always  misuse (\ie over-reliance~\cite{passi2022overreliance}, relying on automation when it performs poorly) or disuse (\ie under-reliance~\cite{yaniv2000advice,bussone2015role,wang2021explanations}, rejecting automated predictions when it is correct) AI systems. 
Such inappropriate reliance results in sub-optimal team performance,  which is always worse than AI alone~\cite{Liu-CSCW-2021,bansal2021does}.
To mitigate such issues, existing work has proposed different interventions including user tutorials~\cite{Lai-CHI-2020,Chiang-IUI-2022}, cognitive force functions~\cite{buccinca2021trust}, and improving AI literacy of the use case~\cite{chiang2021you}. 
Another stream of work proposed to improve the transparency of AI systems with effective explanations~\cite{lai2019human,wang2021explanations}, performance feedback~\cite{Lu-CHI-2021}, and global model properties~\cite{cai2019hello}. 
% ~\glcomment{There are definitely a lot of work. But we don't need to make it very concrete?} 
In summary, these works %~\glcomment{Work or works? I get used to only using ``work'', but I saw works used in context ``a few works''}
presented users with extra information about AI systems (more than advice) or changed users' mindset and knowledge of AI systems. 
% In this work, we proposed to adopt debugging as an intervention to help participants understand the limitations of AI systems and better rely on them.

% \begin{itemize}
%     \item One important goal of human-AI decision making is complementary team performance, which requires appropriate reliance.
%     \item To achieve that goal, users have adopted different interventions like cognitive force functions, user tutorial, AI literacy intervention etc.
%     \item Why these interventions work? They changed users' mindset and understanding of AI system, calibrate the trust. 
%     \item In our paper, we proposed a debugging intervention to help participants understand the limitations of AI system, and better rely on them.
% \end{itemize}

\paratitle{Explanation-based Debugging}. 
% One important stakeholder community of explainable AI~(XAI) is the developers of AI systems. For quality assurance, they need to debug the AI system to improve the robustness of their applications~\cite{preece2018stakeholders}. 
Explanation-based debugging was found to be helpful for improving human understanding of machine learning system~\cite{kulesza2015principles}. Recent works in both natural language processing tasks~\cite{lertvittayakumjorn2021explanation} and computer vision tasks~\cite{balayn2022can} have explored how to leverage explanations for model debugging. 
The core idea of such explanation-based debugging is to check whether the explanations from AI systems misalign with human (expert) knowledge. 
From human feedback, it would be possible to improve machine learning models' robustness, \eg with reducing spurious reasoning patterns~\cite{liu2020towards,sharifi2022should} and bias in dataset~\cite{hu2020crowdsourcing}. Debugging in programming is the process by which programmers can determine the potential errors in the source code and resolve these errors~\cite{turkmen2020investigation}. 
% \gladd{When we put users of AI systems in the shoes of debugging, their mindset changed to doubt whether AI systems make any mistakes, instead of blindly trusting them.} ~\glcomment{To check whether it's appropriate} 
Inspired by such an idea, we proposed debugging as an intervention to help participants understand the limitations of both explanations and advice of AI systems. In such an error finding and resolution process, users may learn when the AI system is trustworthy.
% and conducted empirical work about how explanations will shape user trust and reliance~\cite{wang2021explanations}. 

% \begin{itemize}
%     \item how existing work has touched explanation-based debugging in NLP and CV.
%     \item We get inspiration to take it as an intervention.
% \end{itemize}

Compared with these studies, our focus is to promote appropriate reliance on AI systems by improving users' capability to critically evaluate AI performance at the instance level. For that purpose, we design an elaborate debugging intervention to help users realize the limitations of both AI advice and AI explanation, which may result in calibrated trust in and appropriate reliance on the AI system.%~\glcomment{Do you think the last paragraph redundant with the previous one?} 

%% file: sections/sec-task.tex
% \section{Task, Intervention, and Hypotheses}
\section{Task, Hypotheses, and Intervention}
%\ujcomment{if space needs to be saved, you can get rid of these section intro. lines where you're saying "in this section ..."}
In this section, we describe the deceptive review detection task and present how we designed the  debugging intervention. Based on the explanation-based debugging setting, we further proposed our hypotheses to verify. %\ujcomment{This formulation is incorrect, see comment!}~\glcomment{I changed it a bit, how about now?}~\glcomment{To check, then where we put the explanation-based debugging? Shall we keep it before hypos}
% we describe the deceptive review detection task and present our hypotheses, which have all been preregistered before any data collection.~\glcomment{TO change ``which xxx'', overlap with ANIE / CHI}

\subsection{Deceptive Review Detection Task}
\label{sec:task}

\begin{figure}[h]
    \centering
    \includegraphics[width=0.48\textwidth]{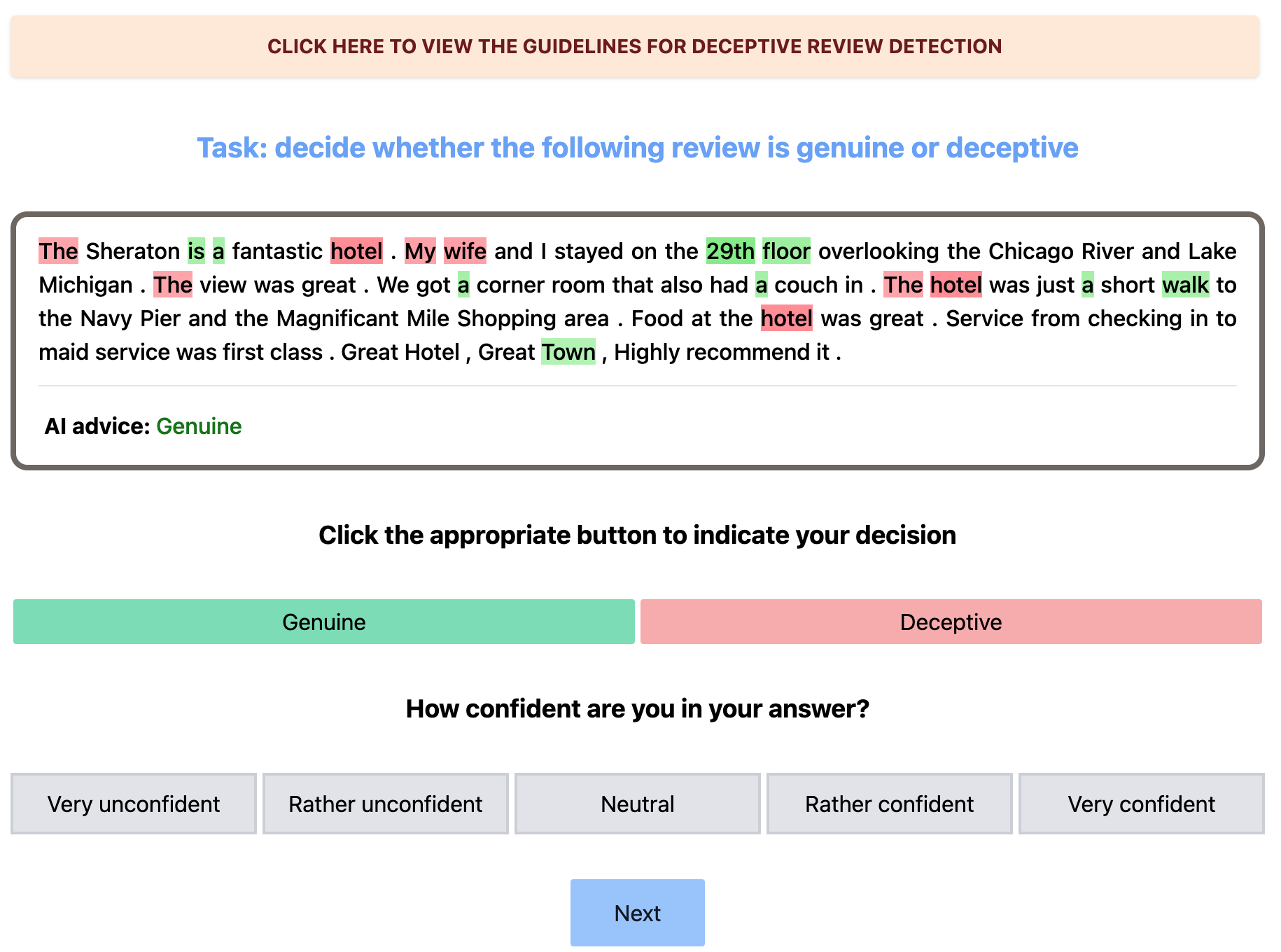}
    \caption{Task interface and an example of the deceptive review detection task.}
    \label{fig:task_interface}
\end{figure}

%~\glcomment{If we have enough space, bring one or two sentences about context} 
%In practice of human-AI decision making, the decision task are typically challenging for human and the AI system achieves superior performance. 
In the context of AI-assisted decision making, the decision tasks are typically challenging for humans, while the AI system may achieve superior performance. 
In this paper, we base our experiment within such a challenging task -- deceptive review detection -- where AI advice can be a realistic need. %\ujcomment{why challenging task?}. 
% We use the publicly available dataset\footnote{\url{https://github.com/vivlai/deception-machine-in-the-loop}}~\cite{lai2019human} of this task. 
In each task, based on a %piece of \ujcomment{see comment about 'a paragraph'}
hotel review, participants are asked to identify whether it is genuine (\ie written by real customers) or deceptive (\ie reviews written by people who did not stay at the hotel). An example of this task is shown in Figure~\ref{fig:task_interface}. 
This task has been used in prior work exploring Human-AI decision making~\cite{Lai-CHI-2020,lai2019human}. We also used the same public dataset~\cite{lai2019human}.\footnote{\url{https://github.com/vivlai/deception-machine-in-the-loop}}
% ~\glcomment{To explain why we choose this task.}
% ~\glcomment{Put a screenshot of interface about task here, and introduce the task.}

% \begin{figure*}[htbp]
%  \centering
%   \subfigure[Decision stage 1.]{\label{fig:task_interface}
%   \centering
%   \includegraphics[height=0.33\textwidth]{figures/decision_stage_1.png}
%  }
%   \subfigure[Decision stage 2.]{\label{fig:helpfulness_ratio}
%   \centering
%   \includegraphics[height=0.33\textwidth]{figures/decision_stage_2.png}
%  }
%  \centering
%  \caption{Screenshots of deceptive review detection task in two stage decision making setting.~\glcomment{To save space, we can only show one (maybe only stage 2)}
%  }
%  \label{fig-two-stage}
% \end{figure*}

\paratitle{Using Text Highlights as Explanations}. In our study, we consider a real-world scenario where the performance of an AI system is not provided or available. To help participants assess the trustworthiness of advice from the AI system in each instance of decision making, we provide local explanations for each prediction. 
Following Lai \etal~\cite{Lai-CHI-2020}, we adopted BERT-LIME (a popular explanation method in text classification tasks) to generate text highlights as local explanations for each AI advice. We first finetuned the BERT~\cite{kenton2019bert} (bert-base-uncased) on the deceptive review detection dataset, and then generated the top-$10$ highlighted features from post-hoc XAI method LIME~\cite{LIME-KDD-2016} as explanations.
% ~\glcomment{@Abri, To check, LIME or SHAP. Which you used} 
% ~\glcomment{Explain why we need this. Based on explanations to interpret trustworthiness}

\paratitle{Selection of Tasks}. 
% ~\glcomment{Do you think data selection would be better. Task selection seems to explain why we select the deceptive review detection task. } 
% Participants were presented with 28 deceptive review detection tasks, of which 8 were training examples and 20 trial cases. 
To measure the effect of the debugging intervention in our study, two batches of tasks with compatible difficulty levels are required. For that purpose, we conducted a pilot study on human performance over 20 tasks randomly sampled from evaluation and test set of the deceptive review detection dataset. We divided the trial cases into two sets of 10 tasks with equal human performance in a pilot study (10 participants). 

\paratitle{Two-stage Decision Making}. 
% In trial cases, participants of all conditions were first presented with a paragraph of hotel review and then asked to make a decision whether that decision is genuine or deceptive. 
% One example of our task interface is shown in Figure~\ref{fig:question-page}. ~\glcomment{Seems the first two sentences are redundant as described previously} 
Following existing empirical study design of human-AI decision making~\cite{green2019principles,green2019disparate}, all participants in our study work on each trial case with two stages of decision making. 
In the first stage, only task input (\ie one paragraph of hotel review) is provided; participants need to make an initial decision on themselves. 
After that, the same task input along with a local explanation (\eg text highlights in review, one example shown in Figure~\ref{fig:task_interface}) and AI advice are provided. They will make the final decision based on all information. To help participants work on this challenging task, we provide a button to access the guidelines in each stage. %~\glcomment{Shall we put this sentence to the experiment setup?}
In addition to making a decision for each task, we also collected participants' confidence in each decision with a 5-point Likert scale: \textit{Very Unconfident}, \textit{Rather Uconfident}, \textit{Neutral}, \textit{Rather Confident}, \textit{Very Confident}.%~\glcomment{We need to make it consistent with the screenshot, we use ``Very'' instead of Completely in our study}
% This first time, they were not presented with the systems' prediction, or with any additional information. After making an initial choice they saw the same case again, but now additionally saw the systems' prediction and text highlight for both labels. 
% ~\glcomment{To check whether we should have one paragraph to explain how the explanations are generated} 
% Participants were then asked to make a final decision. 
% \gladd{This setup of an initial unaided decision and the presentation of advice from an AI system in order to make a second and final choice is similar to the update condition in \cite{green2019principles}, and in line with findings that people first make a decision on their own and only then decide whether to incorporate system advice \cite{green2019disparate}. It also fits with the research of Dietvorst \etal \cite{dietvorst2018overcoming} on trust in two-stage decision-making.}
% ~\glcomment{I copied it from CHI submission and ANIE paper, to update. I think we may simply ignore this orange text in gladd command. We claimed following existing empirical study design.}
% ~\glcomment{I moved ``In all conditions ... shown in Figure'' from section 4.1. Do you think this make experimental conditions easier to follow?}

\subsection{Hypotheses}
% \ujcomment{Rename Section 3 to ``Task and Hypotheses".}
% Some description:
Our experiment was designed to answer questions surrounding the impact of the proposed explanation-based debugging intervention on user estimation of AI performance, and user reliance on AI systems. Putting users into a debugging setting, they will try to challenge the AI advice and explanations. Along with the real-time feedback about the debugging results, they can have a better understanding of how the AI system works and when the explanation and advice are reliable. Thus, they can more accurately estimate the performance of the AI system when no performance of the AI system is provided, and rely on the AI system more appropriately. Based on this, we expect to observe:
\begin{framed}
\textbf{(H1)} Encouraging users to critically evaluate the trustworthiness of AI advice at the instance level in a debugging intervention, will improve their assessment of the AI system's performance at the instance and global levels.
% ~\glcomment{Now, no accuracy is reported, we ask participants to report their estimation of AI accuracy and compare with actual accuracy.}
% Subjecting users to a evaluating the trustworthiness of AI advice will encourage users
% improve their assessment of AI advice trustworthiness.
% training phase will cause them to become more critical of AI predictions.~\glcomment{To discuss, as we don't assess critical mindset in currect analysis} 

\textbf{(H2)} Encouraging users to critically evaluate the trustworthiness of AI advice at the instance level in a debugging intervention, will improve the extent to which users appropriately rely on the system.
% The assessment of trustworthiness of AI advice has a positive effect on the extent to which users appropriately rely on the system.
\end{framed}

%Meanwhile, 
Within a debugging intervention, to present a balanced view of AI systems, we considered showing both the strength and weakness %~\glcomment{Is ``strengths and weaknesses'' correct? Or it's better to use ``strength and weakness''. If so, I will make it consistent in the paper}
of an AI system (by providing accurate or inaccurate advice). 
Thus, multiple tasks of different characteristics %~\glcomment{characteristics or properties}
will be presented in the debugging intervention. When these tasks are presented in different orders, users may show different learning effects, which further affects the reliance on AI systems. Thus, we hypothesize that:
\begin{framed}
\textbf{(H3)} The trustworthiness of AI advice at the instance level in a debugging intervention corresponds to an ordering effect with respect to appropriate reliance.
% The order of tasks in such critical training session has an impact on the final learning effect.
\end{framed}

\subsection{Debugging Intervention}
%\glcomment{To check how we should place this subsection.}
% \glcomment{As debugging is very important method in our study, shall we use a separate subsection for it?}
To help participants accurately assess the trustworthiness of AI advice at the instance level and calibrate their reliance on the AI system, 
% To calibrate users' trust in the AI system and mitigate the over-reliance, 
% we designed a tutorial composed of a brief introductory video and explanation-based debugging phase.
we designed a debugging intervention with explanations generated with post-hoc explanation methods LIME~\cite{LIME-KDD-2016}. Our data and code is available with anonymous companion page.\footnote{\url{https://osf.io/dh34y/?view_only=6a6833eafdbd4d5daa8c036579247159}}
% ~\glcomment{Now we told them AI system will make mistakes, instead of how many mistakes will be made. We will tell participants how many errors will occur}

% \paratitle{Introductory Video}. To help participants better understand how the AI system works, we provide a 3-minute video before the debugging phase. The video explains how machine learning models work generally, and explicate the risk and limitation of such machine learning based AI systems. This also connects with existing work on improving users' machine learning literacy~\cite{Chiang-IUI-2022}. 
% \glcomment{Mention we want to introduce background knowledge, how AI system works, responsibility etc. When describing the knowledge, we can also use terminology ``AI literacy'' or ``machine learning literacy''} 

\begin{figure}[tbp]
    \centering
    \includegraphics[width=0.48\textwidth]{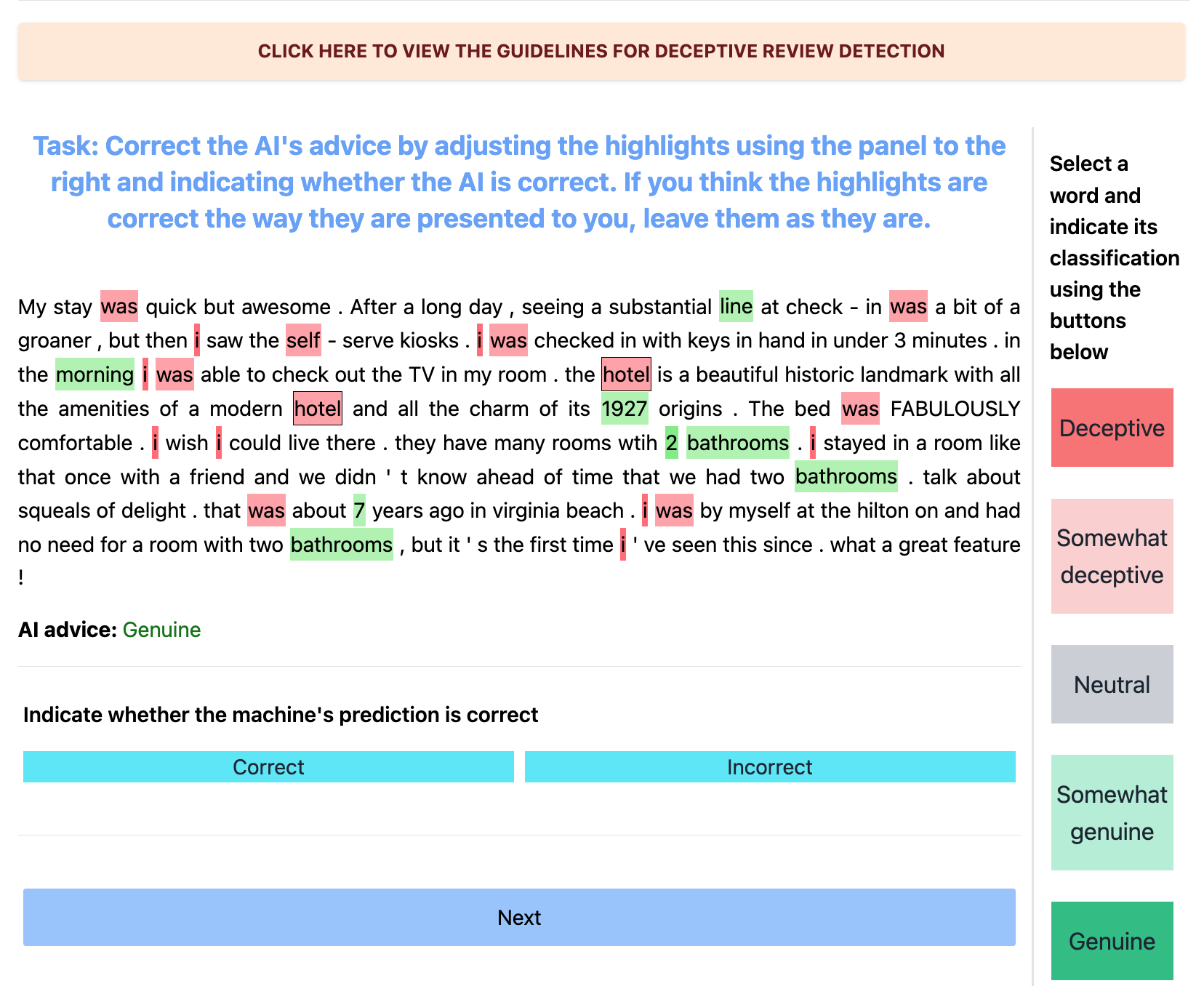}
    \caption{Screenshot of debugging interface.}
    \label{fig:debugging-interface}
\end{figure}

\paratitle{Explanation-Based Human Debugging}. Through the debugging phase, all participants are supposed to learn two important facts about the AI system: (1) the AI advice is not always correct, and (2) explanations are not always informative and helpful in identifying the trustworthiness of AI advice. 
% explanations can be helpful but not always helpful. 
Thus, we considered two main factors for each task: (1) the correctness of AI advice, and (2) whether an explanation is informative (\ie combined with guidelines, whether or not such explanations can help participants easily identify the correct answer). 
% (at least one highlighted token aligned with task guidelines and may help infer the trustworthiness of AI advice) 
% In the debugging phase, to ensure explanations can support users in identifying trustworthy AI advice, we need informative explanations, which can help participants identify the correct answers according to task guidelines. 
Participants subjected to training were presented with a hotel review with explanatory elements consisting of a model prediction and color-coded highlights showing 10 predominant features. Each token highlight shows the contribution of the token to the model prediction on a 5-point Likert scale: \textit{deceptive}, \textit{somwhat deceptive}, \textit{neutral},  \textit{somewhat genuine}, \textit{genuine}. 
% \textit{somewhat genuine}, \textit{neutral}, \textit{somwhat deceptive}, \textit{deceptive}. 
This difference in the contribution is distinguished by the color and intensity of the highlight shown in the interface. An example of the debugging phase is shown in Figure~\ref{fig:debugging-interface}. 
They are instructed to read the text, and, when deemed necessary, refine the explanations by adjusting the highlights and indicating whether the AI advice is correct. After each task, the correctness of AI advice and missed adjustments will be shown to the participant as real-time feedback. 
In practice, the explanations obtained from XAI methods may not always align with human understanding~\cite{sokol2020one}. 
Besides realizing the explanations are not always helpful, we hope participants can learn patterns they can rely on to make the decision given the guidelines. With that wish, the authors manually adjusted the highlights generated with BERT-LIME according to the task guidelines (from~\cite{Lai-CHI-2020}) and take the adjusted highlights as ground truth for debugging phase. 
% Taking the adjusted highlights as ground truth, all participants were presented with the explanations generated with BERT-LIME top-$10$ features.

\begin{figure}[h]
    \centering
    \includegraphics[width=0.48\textwidth]{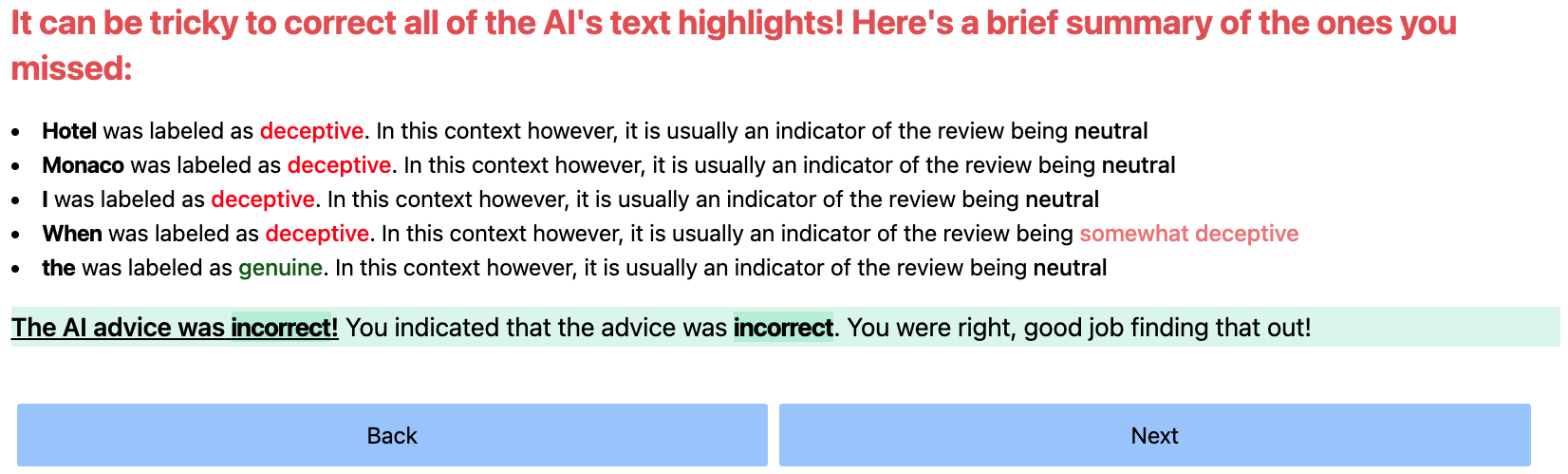}
    \caption{Screenshot of debugging feedback.}
    \label{fig:feedback}
\end{figure}

\paratitle{Real-time Feedback in Debugging Intervention}. We provide the real-time feedback in each debugging task, to show whether AI advice is correct and which highlights participants missed to adjust according to guidelines. One example of the feedback is shown in Figure~\ref{fig:feedback}.

% ~\glcomment{Describe the main factors considered: (1) Whether AI advice is correct (2) Whether explanation is meaningful} 
\paratitle{{Selection of the Debugging Tasks}}. To create a balance between the strength and weakness of the AI system, we manually selected four tasks with informative explanations (where explanations and guidelines can help participants easily identify the correct answer) and four tasks with uninformative  explanations. 
The eight tasks presented in our debugging phase are: (1) two tasks with correct AI advice and informative explanations, (2) two tasks with correct AI advice and uninformative explanations, (3) two tasks with incorrect AI advice and informative explanations, (4) two tasks with incorrect AI advice and uninformative explanations. The tasks are balanced in whether explanations are informative and whether AI advice is correct. 
While the informative explanations are manually selected, the correctness of AI advice is determined randomly.

\paratitle{Ordering Effect}. When presenting the debugging phase to participants, the order of tasks may have an impact on their estimation of AI performance and reliance on the AI system. According to existing work~\cite{Tolmeijer-UMAP-2021,nourani2021anchoring}, first impressions (either good or bad) greatly affect user estimation of AI performance and user trust in AI systems. 
% Meanwhile, the curriculum designs were also proved to be effective in improving critical mindset / critical thinking skills. 
Overall, both correct AI advice and informative explanations tend to leave positive impression on users. 
As pointed out by a recent study~\cite{nussberger2022public}, the public would prioritize the accuracy of AI systems over interpretability. 
% As users typically outweigh the correctness of AI advice, the correctness of AI advice may be more important in the context of human-AI decision making. 
% When comparing the two cases, the impact of correct AI advice may be more direct. 
Thus, compared with ``wrong AI advice, informative explanation'' case, we would consider ``correct AI advice, uninformative explanation'' will leave participants a better impression. With these concerns, we designed three orders of tasks:
% ~\glcomment{To discuss, whether we should use ``meaningful'' for explanation, and whether we should use ``difficulty'' to characterize the order features.}
\begin{itemize}
    \item Random order.
    \item Decreasing impression order (\ie from good to bad): correct AI advice, informative explanation $\rightarrow$ correct AI advice, uninformative explanation $\rightarrow$ wrong AI advice, informative explanation $\rightarrow$ wrong AI advice, uninformative explanation.
    % ~\glcomment{correct advice first}
    \item Increasing impression order (\ie from bad to good): wrong AI advice, uninformative explanation $\rightarrow$ wrong AI advice, informative explanation $\rightarrow$ correct AI advice, uninformative explanation $\rightarrow$ correct AI advice, informative explanation.
    % ~\glcomment{incorrect AI advice first}
    % \item \glcomment{Understanding whether there is a perception of learning AI system?}
\end{itemize}

%% file: sections/sec-exp.tex
\section{Study Design}
This section describes our experimental conditions, variables, participants, and procedure in our study. {This  study was approved by the human research ethics committee of our institution.} More implementation details can be found in the appendix (\ref{sec-appendix-implementation}).
% ~\glcomment{Remember to mention our study has been approved by institution agent.}
% ~\glcomment{Check overlap between sec 3 and sec 4. It's now a placeholder, change the content of these subsections.}
% ~\glcomment{Now, our comparison is mainly on the debugging phase on trust calibration? Shall we also assess participants' performed reliability of AI system or sth? Then our main focus shift to whether trust calibration will reduce over-reliance and promote appropriate reliance? }

\subsection{Experimental Conditions}

In our study, all participants worked on deceptive review detection tasks with a two-stage decision making process (described in Sec.~\ref{sec:task}). In all conditions, the top-$10$ most important features obtained from BERT-LIME are highlighted as an explanation for AI advice to help participants identify the trustworthiness of AI advice. 
% ~\glcomment{If we decide not to use tutorial in description, we should change the condition name in command}

\begin{table*}[htbp]
	\centering
	\caption{The different variables considered in our experimental study. ``DV'' refers to the dependent variable.}
	\label{tab:variables}
	\begin{small}
	\begin{tabular}{c | c | c | c}
	    \hline
	    \textbf{Variable Type}&	\textbf{Variable Name}& \textbf{Value Type}& \textbf{Value Scale}\\
	    \hline \hline
        \multirow{5}{*}{Assessment (DV)}& EAP& Continuous, Interval& [0, 10]\\
        & ETP& Continuous, Interval& [0, 10]\\
        & MAP& Continuous, Interval& [0, 10]\\
        & MTP& Continuous, Interval& [0, 10]\\
        & CCD& Continuous, Interval& [0, 10]\\
	    \hline
	    \multirow{4}{*}{Reliance (DV)}& Agreement Fraction& Continuous, Interval& [0.0, 1.0]\\
	    & Switch Fraction& Continuous, Interval& [0.0, 1.0]\\
	    & RAIR& Continuous, Interval& [0.0, 1.0]\\
	    & RSR& Continuous, Interval& [0.0, 1.0]\\
	    \hline
	    Performance (DV)& Accuracy& Continuous, Interval& [0.0, 1.0]\\
	    \hline
     \multirow{4}{*}{Trust (DV)}& TiA-R/C& Likert& 5-point, 1: poor, 5: very good\\
	 & TiA-U/P& Likert& 5-point, 1: poor, 5: very good\\
	 & TiA-IoD& Likert& 5-point, 1: poor, 5: very good\\
	 & TiA-Trust& Likert& 5-point, 1:strong distrust, 5: strong trust\\
     \hline

     \multirow{3}{*}{Covariates}& ATI& Likert& 6-point, 1: low, 6: high\\
     & TiA-PtT& Likert& 5-point, 1: tend to distrust, 5: tend to trust \\
     & TiA-Familiarity& Likert& 5-point, 1: not familiar, 5: very familiar \\
    %  \hline
	   % Covariates & Comparison to Peers& Continuous, Interval& [0,100]\\ 
        % & DKE& Continuous, Interval& [-6,6]\\
	    % & Self estimation& Likert& 7-point, 0: none answer right, 6: all answered right\\
	    % & Estimation of peers& Likert& 7-point, 0: none answer right, 6: all answered right\\
	   % \cline{2-4}
	    % \hline
	   % \multirow{2}{*}{Other}& Experimental Condition & Categorical& \{w/o Tutorial\} $\times$ \{w/o XAI\}\\
	    % &Experimental Condition & Categorical& \{XAI\}\\
	    %\hline
	    %\multirow{2}{*}{\tabincell{c}{Exploratory \\Demographic Variable}}& Gender& Categorical& \{Male, Female, Other\}\\
	    %& Age& Continuous, Interval& [18.0, 200.0]\\
	   % \hline
	   % Other& Helpfulness of Explanation& Likert& 5-point, 1: not helpful, 5: very helpful\\
	    \hline
	\end{tabular}
	\end{small}
\end{table*}

The differences between conditions are whether debugging intervention is adopted and the order of debugging tasks. 
% whether intervention / tutorial is presented and whether explanations are provided along with AI advice. 
To comprehensively study the effect of debugging intervention, we considered four experimental conditions in our study: (1) no debugging intervention (represented as \control{}), 
% ~\glcomment{Question: shall we also adopt random, increasing, decreasing order in control condition? Do we want to make ablation with the video part? Try to show video, and video + debugging has different impact?} 
(2) with debugging intervention, debugging tasks in random order (represented as \tutorialrandom{}), (3) with debugging intervention, debugging tasks in decreasing impression order (represented as \tutorialdec{}), (4) with debugging intervention, debugging tasks in increasing impression order (represented as \tutorialinc{}).
In conditions with debugging intervention, participants were presented with eight selected tasks with performance feedback and manually adjusted contribution of tokens. 
While in \control{} condition, the eight tasks selected are presented as normal tasks without any feedback of AI advice correctness or adjusted explanation feedback. Such a control setting is designed to compare with debugging intervention and eliminate the learning effect brought by the eight tasks. 

For each batch of ten tasks, {the AI system} was configured to provide correct advice on eight of them and incorrect advice on two tasks. 
% So the accuracy of AI systems is $80\%$. 
To eliminate the potential ordering effect of trial cases, we randomly assigned one batch of selected tasks (see section~\ref{sec:task}) as the first batch and further shuffled the task order within each batch. 
% To avoid any ordering effect brought by trial cases, we randomly assign one batch of tasks as first batch of tasks for each participant and further shuffled the order of tasks within each batch.~\glcomment{Before final submission, rephrase this subsection, to avoid largely overlap with CHi submission}

\subsection{Measures And Variables}
To have a more comprehensive view of variables used in our experimental analysis, we listed the main variables in Table~\ref{tab:variables}. 
Notice that we do not add the confidence and dimensions from the NASA-TLX questionnaire ~\cite{NASA-TLX-1988} into it.

% Table~\ref{tab:variables} presents an overview of all the variables considered in our study.~\glcomment{@Abri, You can replace all variables in the Table~\ref{tab:variables} with what we're going to use}
% ~\glcomment{Current description copied from CHI submission, to modify before final submission}
To verify \textbf{H1}, we assessed participants' global estimation of AI system's performance with two questions: ``From the previous 10 tasks, on how many tasks do you estimate the AI advice to be correct?'' and ``From the previous 10 tasks, how many questions do you estimate to have been answered correctly? (after receiving AI advice)''. 
The answers to the two questions correspond to participants' estimation of AI performance and team performance respectively. %~\glcomment{Do you think we should claim it's self-performance estimation instead of team performance} 
{We can refer to the estimated trustworthiness as estimated AI performance (\textbf{EAP}) and estimated team performance (\textbf{ETP}).}
Comparing that performance estimation with actual performance in abstract difference, we can calculate the degree of miscalibration of AI performance (\textbf{MAP}) and team performance (\textbf{MTP}). 
% with the abstract value of the difference between estimated correct number and actual correct number. 
If participants can accurately estimate the performance of AI system at instance level, they may make the final decision with high confidence. Thus, for the AI performance estimation at instance level, we calculated the number of tasks they made the correct final decision with indication of ``Very Confident'' (\textbf{CCD}).
% ~\glcomment{Is Confident Correct Decisions appropriate?}

\begin{table}[htbp]
	\centering
	\caption{The different appropriate reliance patterns considered in~\cite{schemmer2022should}. $d_i$ and $d_f$ refer to initial human decision and final human decision respectively. $\checkmark$ and $\times$ refer to correct and incorrect respectively. }
	\label{tab:reliance_patterns}
	\begin{tabular}{c | c | c | c}
	    \hline
        \textbf{$d_i$} & \textbf{AI advice} & \textbf{$d_f$} & \textbf{Reliance}\\
        \hline
        \hline
        $\times$& $\checkmark$& $\checkmark$& Positive AI reliance\\
        $\times$& $\checkmark$& $\times$& Negative self-reliance\\
        $\checkmark$& $\times$& $\checkmark$& Positive self-reliance\\
        $\checkmark$& $\times$& $\times$& Negative AI reliance\\
    \hline
	\end{tabular}
\end{table}

To verify \textbf{H2} and \textbf{H3}, we measured both reliance and appropriate reliance of participants on the AI system. The reliance is measured with two widely adopted metrics: the \textbf{Agreement Fraction} and the \textbf{Switch Fraction}. These look at the degree to which participants are in agreement with AI advice, and how often they adopt AI advice in cases of initial disagreement. They are commonly used in the literature, for example in \cite{yin2019understanding, zhang2020effect,Lu-CHI-2021}. As for the appropriate reliance, we followed Max \etal~\cite{schemmer2022should} to calculate the appropriate reliance based on four reliance patterns (shown in Table~\ref{tab:reliance_patterns}). 
According to the four reliance patterns where human initial decision disagree with AI advice and the correct answer occurs in one of them, we can assess the appropriate reliance from two dimensions:
$$\textnormal{\textbf{RAIR}} = \frac{\textnormal{Positive AI reliance}}{\textnormal{Positive AI reliance + Negative self-reliance}},$$
$$\textnormal{\textbf{RSR}} = \frac{\textnormal{Positive self-reliance}}{\textnormal{Positive self-reliance + Negative AI reliance}}.$$
They stand for whether users switch to AI advice when AI outperforms them, and whether users can insist on correct decisions made by themselves when AI advice is incorrect. In addition, we consider the accuracy in batches to measure participants' performance with AI assistance. 

 \begin{figure*}[htbp]
    \small
    \centering
    \includegraphics[scale=0.4]{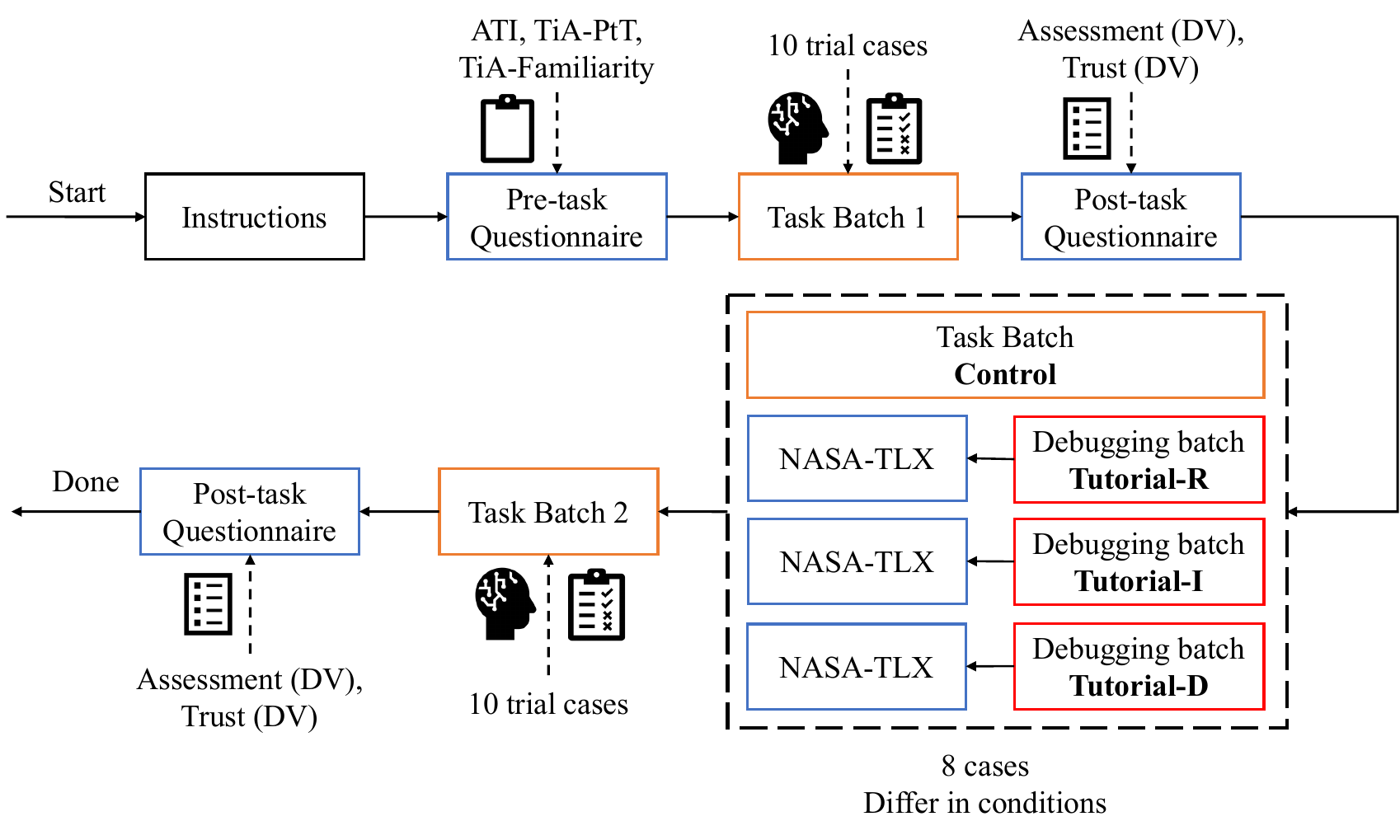}
    \caption{Illustration of the procedure that participants followed within our study. Blue boxes represent questionnaire phase, orange boxes represent task phase, and the red box represents the debugging intervention.}%~\glcomment{To improve further, if we don't have enough time, keep it as is.}}
    \label{fig:procedure}
\end{figure*}

For a deeper analysis of our results, a number of additional measures were considered based on observations from existing literature~\cite{schramowski2020making,li2019no,Tolmeijer-UMAP-2021}:
% ~\ujcomment{add some refs. where studies considered these measures and found effects}:
\begin{itemize}
    \item Trust in Automation (TiA) questionnaire \cite{Korber-2018-TiA}, a validated instrument to measure (subjective) trust \cite{Tolmeijer-UMAP-2021} consisting of 6 subscales: \textit{Reliability/Competence}~(TiA-R/C), \textit{Understanding/Predictability}~(TiA-U/P), \textit{Propensity to Trust}~(TiA-PtT), \textit{Familiarity}~(TiA-Familiarity), \textit{Intention of Developers} (TiA-IoD), and \textit{Trust in Automation}~(TiA-Trust). 
    \item Affinity for Technology Interaction Scale (ATI)~\cite{Franke-2019-ATI}, administered in the pre-task questionnaire. Thus, we account for the effect of participants' affinity with technology on their reliance on systems ~\cite{Tolmeijer-UMAP-2021}.
    \item NASA-TLX questionnaire~\cite{NASA-TLX-1988} for the working load assessment of the debugging intervention.
\end{itemize}

\subsection{Participants}

\paratitle{Sample Size Estimation}. Before recruiting participants, we computed the required sample
size in a power analysis for a Between-Subjects ANOVA using G*Power~\cite{faul2009statistical}. 
{To correct for testing multiple hypotheses, we applied a Bonferroni correction so that the significance threshold decreased to $\frac{0.05}{3}=0.017$.} 
We specified the default effect size $f = 0.25$
(\textit{i.e.,} indicating a moderate effect), a significance threshold $\alpha = 0.017$ (\textit{i.e.,} due to testing multiple hypotheses), a statistical power of $(1 - \beta) = 0.8$, and that we will investigate $4$ different experimental conditions. This resulted in a required sample size of $230$ participants. We thereby recruited 324 participants from the  crowdsourcing platform Prolific\footnote{\url{https://www.prolific.co}}, in order to accommodate potential exclusion.

\paratitle{Compensation}. All participants were rewarded with \pounds 3.8, amounting to an hourly wage of \pounds 7.6 (estimated completion time was 30 minutes).
% \stcomment{This makes for 1.8 * 6 = 8.1 pounds hourly wage. If I do 1.8/7.5 then I get to estimated completion time of 14.4 minutes}~\glcomment{I checked it again, actually we run two rounds for main studt. First with estimation of 12 minutes (\pound 1.5), and second with 10 minutes (\pound 1.25). I think we can write it as 1.5. I included the service fee and calculated the mean for all participants.}
We rewarded participants with extra bonuses of \pounds 0.05 for every correct decision in the 20 trial cases. 
Such extra bonus for correct decisions provides a monetary motivation for crowd workers to try their best on each task, which is also widely adopted by existing work~\cite{chiang2021you,Lai-CHI-2020}.
% By incentivizing participants to reach a correct decision, we operationalize the concomitant ``vulnerability'' discussed by Lee and See\cite{lee2004trust} as a contextual requirement to encourage appropriate system reliance. 
% ~\glcomment{To reduce overlap with previous submissions}

\paratitle{Filter Criteria}. All participants were proficient English speakers above the age of 18. For a high-quality study, we require participants to have an approval rate of at least 90\% and more than 80 successful submissions on the Prolific platform. After reading the basic introduction and guidelines about the deceptive review detection task, participants who failed any qualification test (about understanding the task) were removed from our study. 
After data collection, we excluded participants from our analysis if they failed any 
attention check (90 participants). The resulting sample of 234 participants had an average age of 39 ($SD=13$) and a gender distribution ($48.7\%$ female, $49.6\%$ male, $ 1.7\%$ other). 
% ~\glcomment{Remember to have qualification questions and attention checks among tasks.} 

\subsection{Procedure}
\label{sec:procedure}
The full procedure of our study can be visualized in Figure~\ref{fig:procedure}. 
In the beginning, all participants will be presented with a basic introduction of the deceptive review detection task. According to Lai \etal~\cite{Lai-CHI-2020}, guidelines about how to identify deceptive reviews are highly useful in improving user performance on this task. Thus, we also follow them to provide the guidelines in the introduction. 
Then, participants will be checked with two qualification questions to ensure they carefully read the instruction and understand this task. 
Any failure at the qualification test will result in removal from our study. All reserved participants will then be asked to answer a pre-task questionnaire consisting of affinity for technology interaction, TiA-PtT, and TiA-Familiarity. 

%  \begin{figure*}[htbp]
%     \small
%     \centering
%     \includegraphics[scale=0.4]{figures/procedure.pdf}
%     \caption{Illustration of the procedure that participants followed within our study. Blue boxes represent questionnaire phase, orange boxes represent task phase, and the red box represents the debugging intervention.}%~\glcomment{To improve further, if we don't have enough time, keep it as is.}}
%     \label{fig:procedure}
% \end{figure*}

As described in section~\ref{sec:task}, we selected two batches of tasks (10 for each batch) as trial cases and 8 tasks for debugging intervention. For all conditions, participants will first work on the first batch of tasks and go through a post-task questionnaire for assessment of AI performance and subjective trust in AI system (\ie with TiA subscales). 
% ~\glcomment{To check whether trustworthiness is the best terminology in our study} 
The main difference between conditions (shown in the dashed box of Figure~\ref{fig:procedure}) is the 8 tasks presented after the post-task questionnaire. 
In condition \control{}, participants will work on the 8 tasks as normal trial cases. No debugging intervention and result feedback will be provided. 
In comparison, we show debugging intervention and result feedback in conditions \tutorialrandom{}, \tutorialinc{}, and \tutorialdec{}. 
% And for the three conditions with debugging intervention
In conditions with debugging intervention, the participants will go through the debugging tasks with different task orders and be asked about the task working load resulting from the debugging intervention, using the NASA-TLX~\cite{NASA-TLX-1988} questionnaire. 
% After the tutorial phase, we assessed the cognitive load with NASA-TLX~\cite{NASA-TLX-1988} questionnaire for all participants who took the tutorial. 
Then, participants in all conditions will continue to work on another batch of tasks and answer the same post-task questionnaire as the one after the first task batch. 

%% file: sections/sec-result.tex
\section{Results and Analysis}
In this section, we present the main results of our study (\ie hypothesis tests) and further exploration about reliance shaping with confidence dynamics. 

\subsection{Descriptive Statistics}
In our analysis, we only consider participants who passed all attention checks, as a measure of participant reliability~\cite{gadiraju2015understanding}.
Participants were distributed in a balanced fashion over experimental conditions: 57 (\control{}), 59 (\tutorialrandom{}), 60 (\tutorialdec{}), 58 (\tutorialinc{}). 
On average, participants spend around 51 minutes ($SD = 14$) in our study. 
% With Kruskal Wallis test, we compared the time spent by participants of different conditions. The time spent on conditions with debugging intervention has no significant difference. However, we found that participants in condition \tutorialdec{} and \tutorialinc{} spent significantly more time than participants in condition \control.

\paratitle{Variable Distribution}. The covariates' distribution is as follows: \textit{ATI} ($M = 3.91$, $SD = 0.94$, 6-point Likert scale, \textit{1: low, 6: high}), \textit{TiA-PtT} ($M = 2.89$, $SD = 0.61$, 5-point Likert scale, \textit{1: tend to distrust}, \textit{5: tend to trust}), \textit{TiA-Familiarity} ($M = 2.29$, $SD = 1.09$, 5-point Likert scale, \textit{1: unfamiliar with AI system used in study}, \textit{5: familiar with AI system used in study}).

\begin{figure}[htbp]
    \centering
    \includegraphics[scale=0.28]{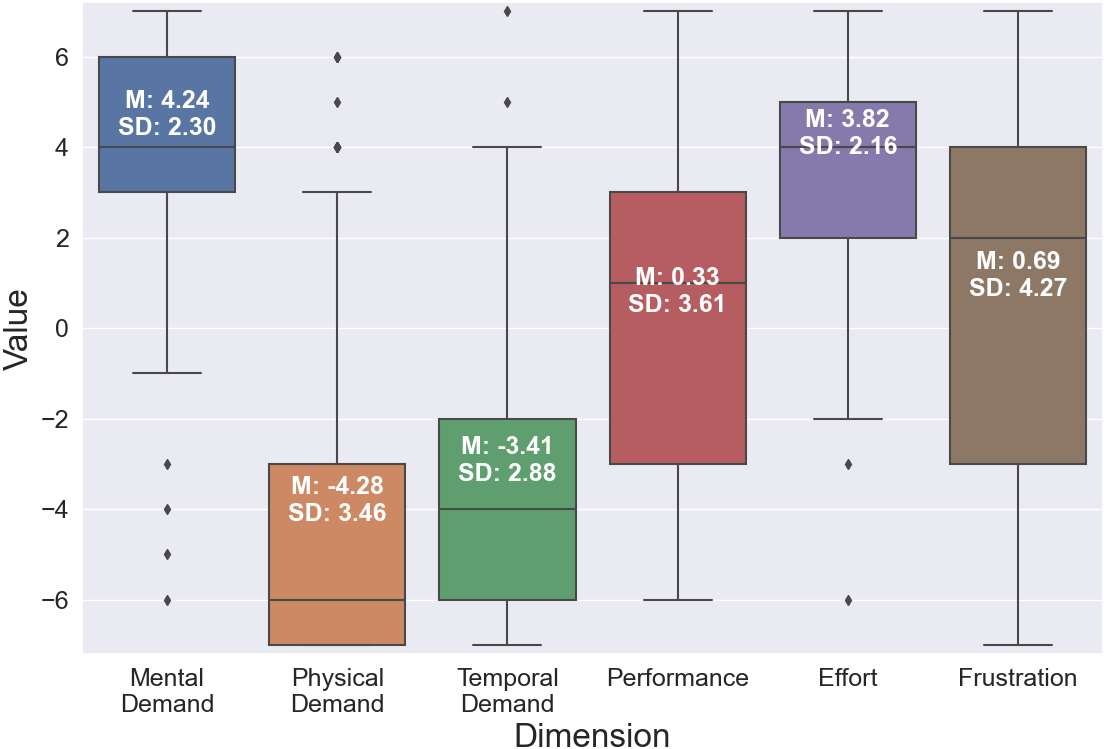}
    \caption{Box plot illustrating the distribution of the different dimensions in NASA-TLX questionnaire. $M$ and $SD$ represent mean and standard deviation respectively.}
    \label{fig:nasa_tlx}
\end{figure}

The working load of debugging intervention is measured with NASA-TLX questionnaire (scale in [-7, 7]). For all dimensions except ``Performance'', a higher value indicates a higher working load. In the dimension ``Performance'', a smaller value indicates a higher estimated performance on tasks. We visualized the dimensions in Figure~\ref{fig:nasa_tlx}. 
In general, participants think the debugging intervention requires high ``Mental Demand'' and ``Effort'', but low ``Physical Demand'' and ``Temporal Demand''. While most participants do not show high expectations in achieved ``Performance'', they also do not get troubled with ``Frustration''.

\paratitle{Performance Overview}. On average across all conditions, participants achieved an accuracy of $0.64$ ($SD = 0.11$) over the two batches of tasks, still lower than the aforementioned AI accuracy of $0.8$. The agreement fraction is $0.66$ ($SD = 0.13$) while the switching fraction is $0.31$ ($SD = 0.22$). With these measures, we confirm that when disagreement appears participants in our study did not always switch to AI advice and participants did not blindly rely on the AI system. 
In the two batches of tasks (10 for each batch), the average estimated AI performance are 5.81 ($SD=1.91$) and 5.79 ($SD=1.71$) respectively; the average estimated team performance is 6.64 ($SD=1.74$) and 6.44 ($SD=1.87$) respectively. Overall, participants underestimated the performance of the AI system and believed they could outperform the AI system on this task after receiving AI advice.

\subsection{Hypothesis Tests}
\subsubsection{H1: the effect of critical evaluation setting on AI performance estimation}

To verify H1, we used Wilcoxon signed rank tests to compare all assessment-based dependent variables of participants before and after the debugging intervention (only participants in condition \tutorialrandom{}, \tutorialdec{}, \tutorialinc{} are considered). The results are shown in Table~\ref{tab:hypothesis-res-1}. %\gladd
{Although no significant results were found to support \textbf{H1}, we found that participants in \tutorialdec{} condition showed a worse \textbf{MTP} after the debugging intervention, in contrast to our expectations.}
% \gladd{No significant difference exists in the assessment-based dependent variables after the debugging intervention.} 
% Only on condition \tutorialdec{}, participants showed a significantly different estimation of team performance. Post-hoc Mann-Whitney tests using a Bonferroni-adjusted alpha level of 0.017 were used to make pairwise comparisons of performance, revealing that after receiving debugging intervention with decreasing impression order, the miscalibration of team performance becomes even more severe, which is against our \textbf{H1}. 
Thus, \textbf{H1} is not supported.
% ~\glcomment{Consider abbreviation for saving space} 
% Only participants in condition \tutorialrandom{}, \tutorialinc{}, \tutorialdec{} are considered in analysis. 

\begin{table}[htbp]
	\centering
	\caption{Wilcoxon signed ranks test results for \textbf{H1} on AI performance estimation. ``$\dagger$'' indicates the effect of variable is significant at the level of 0.017 (adjusted alpha).}%\glcomment{According to our CHI paper, I think we should use one-side assumption in wilcoxon test. Thus the results are updated again}}
 % ``$\dagger$'' indicates the effect of variable is significant at the level of 0.017 (adjusted alpha).
	\label{tab:hypothesis-res-1}%
	\begin{footnotesize}
	\begin{tabular}{c | c c | c c| c c | c c}
	    \hline
	    \textbf{Condition}&	\multicolumn{2}{c|}{\textbf{Debugging}}&	\multicolumn{2}{c|}{\tutorialrandom{}}&	\multicolumn{2}{c|}{\tutorialdec{}}&	\multicolumn{2}{c}{\tutorialinc{}} \\
	    \hline
	    \textbf{DV}& $T$& $p$& $T$& $p$& $T$& $p$& $T$& $p$\\
	    \hline \hline
	    \textbf{MAP}& 3833& .662& 363& .742& 463& .238&457 &.826\\
	    % \textbf{MTP}& 3716& .090& 525& .373& 231& \textbf{.005}$^{\dagger}$& 365&.261\\
     \textbf{MTP}& 4006& .957& 512& .892& 324& \textbf{.992}$^\dagger$& 528&.160\\
	    \textbf{CCD}& 3761& .753& 474& .717& 379& .660&429 &.603\\
	    \hline
	\end{tabular}%
	\end{footnotesize}
\end{table}

\begin{table*}[htbp]
	\centering
	\caption{Participants' estimation of AI performance and Team Performance.}
	\label{tab:estimation}%
	% \begin{footnotesize}
	\begin{tabular}{c | c c |c c | c c}
	    \hline
	    \textbf{Condition}&	\multicolumn{2}{c|}{\tutorialrandom{}}&	\multicolumn{2}{c|}{\tutorialdec{}} &	\multicolumn{2}{c}{\tutorialinc{}}\\
	    \hline
	    \textbf{Estimation}& Before& After& Before& After& Before& After\\
	    \hline \hline
     \textbf{EAP}& $6.05 \pm 1.63$& $5.97 \pm 1.65$& $5.92 \pm 1.89$& $6.07 \pm 1.67$& $6.00 \pm 2.20$& $5.64 \pm 1.90$\\
	    % \textbf{AI Performance}& $6.05 \pm 1.63$& & $6.00 \pm 2.20$& $5.97 \pm 1.65$& $6.07 \pm 1.67$&$5.64 \pm 1.90$\\
      \textbf{ETP}& $6.81 \pm 1.55$& $6.36 \pm 1.85$& $6.68 \pm 1.48$& $6.57 \pm 1.65$& $6.81 \pm 1.90$& $6.60 \pm 1.92$\\
        % \textbf{Team Performance}& $6.81 \pm 1.55$& $6.68 \pm 1.48$& $6.81 \pm 1.90$& $6.36 \pm 1.85$& $6.57 \pm 1.65$&$6.60 \pm 1.92$\\
	    % \textbf{CCD}& 3761& .753& 474& .717& 379& .660&429 &.603\\
	    \hline
	\end{tabular}%
	% \end{footnotesize}
\end{table*}

\begin{table*}[htbp]
	\centering
	\caption{Wilcoxon signed ranks test results for \textbf{H2} on reliance-based dependent variables. ``$\dagger$'' indicates the effect of variable is significant at the level of 0.017 (adjusted alpha).}
	\label{tab:hypothesis-res-2}%
% 	\begin{small}
	\begin{tabular}{c | c  c | c c | c c | c c}
	    \hline
	    \textbf{Condition}&	\multicolumn{2}{c|}{\textbf{Debugging}}&	\multicolumn{2}{c|}{\tutorialrandom{}}&	\multicolumn{2}{c|}{\tutorialdec{}}&	\multicolumn{2}{c}{\tutorialinc{}} \\
	    \hline
	    \textbf{Dependent Variables}& $T$& $p$ & $T$& $p$& $T$& $p$ & $T$& $p$\\
	    \hline \hline
	    \textbf{Accuracy}& 6494& \textbf{.998}$^\dagger$ & 840& .952& 679& .935& 684&.970 \\
	    \textbf{Agreement Fraction}& 6332& .896 &756& .639 & 512& .475& 897& .971\\
	    \textbf{Switch Fraction}& 6812& .953 & 703& .735 &762 & .565 & 817& .979\\
	    \textbf{RAIR}& 6340& .981 & 628& .829 & 722& .618 & 807&\textbf{.995}$^\dagger$\\
	    \textbf{RSR}& 2494& .736& 241 & .325 & 311& .953 & 292&.461\\
	    \hline
	\end{tabular}%
% 	\end{small}
\end{table*}

{To have a closer look at how participants' assessment of the AI performance and Team performance change after debugging intervention. 
We compared the assessment of AI performance and team performance across conditions. With Kruskal-Wallis H-test, we found no significant difference in the estimation across conditions with debugging intervention.  
We show their estimation with mean value and standard deviation ($M \pm SD$) in Table~\ref{tab:estimation}. We found that (1) generally, participants showed a worse estimation of AI performance and team performance after the debugging intervention; (2) only participants in the \tutorialdec{} condition showed a slight increase in the estimation of AI performance.}

\subsubsection{H2: the effect of critical evaluation setting on appropriate reliance}
Similarly, to analyze the effect of the debugging intervention on user reliance on the AI system (\textbf{H2}), we used Wilcoxon signed rank tests to compare all reliance-based dependent variables of participants before and after the debugging intervention. The results are shown in Table~\ref{tab:hypothesis-res-2}. 
{Overall in all conditions with the debugging intervention, the improvement in reliance caused by debugging intervention was not statistically significant. 
With a post-hoc Mann-Whitney test on \textbf{Accuracy}, we found that: after the debugging intervention, the accuracy drops significantly. For a fine-grained analysis, we further conducted Wilcoxon signed rank tests on each condition with the debugging intervention. 
We found that participants in the \tutorialinc{} condition show a significant difference in \textbf{RAIR},  while no significant difference is found with post-hoc Mann-Whitney test.
}
% Overall, participants showed significantly less reliance after receiving debugging interventions (confirmed with post-hoc Mann-Whitney tests). 
% For a fine-grained analysis, we further conducted Wilcoxon signed rank tests on each condition with debugging intervention. 
% As it shows, participants in \tutorialinc{} condition show a significant difference in reliance-based measures while no significant difference is found in condition \tutorialdec{} and \tutorialrandom{}. With post-hoc analysis, we found that participants in \tutorialinc{} condition showed significantly less reliance and appropriate reliance on AI system after debugging intervention. 
The observed results do not support the \textbf{H2}.

% \begin{table*}[htbp]
% 	\centering
% 	\caption{Wilcoxon signed ranks test results for \textbf{H2} on reliance-based dependent variables. ``$\dagger$'' indicates the effect of variable is significant at the level of 0.017 (adjusted alpha).}
% 	\label{tab:hypothesis-res-2}%
% % 	\begin{small}
% 	\begin{tabular}{c | c  c | c c | c c | c c}
% 	    \hline
% 	    \textbf{Condition}&	\multicolumn{2}{c|}{\textbf{Debugging}}&	\multicolumn{2}{c|}{\tutorialrandom{}}&	\multicolumn{2}{c|}{\tutorialdec{}}&	\multicolumn{2}{c}{\tutorialinc{}} \\
% 	    \hline
% 	    \textbf{Dependent Variables}& $T$& $p$ & $T$& $p$& $T$& $p$ & $T$& $p$\\
% 	    \hline \hline
% 	    \textbf{Accuracy}& 6494& \textbf{.998}$^\dagger$ & 840& .952& 679& .935& 684&.970 \\
% 	    \textbf{Agreement Fraction}& 6332& .896 &756& .639 & 512& .475& 897& .971\\
% 	    \textbf{Switch Fraction}& 6812& .953 & 703& .735 &762 & .565 & 817& .979\\
% 	    \textbf{RAIR}& 6340& .981 & 628& .829 & 722& .618 & 807&\textbf{.995}$^\dagger$\\
% 	    \textbf{RSR}& 2494& .736& 241 & .325 & 311& .953 & 292&.461\\
% 	    \hline
% 	\end{tabular}%
% % 	\end{small}
% \end{table*}

{Although no significant improvement was found in the performance and reliance measures due to debugging intervention, we did witness a drop in reliance measures generally: \textbf{Accuracy} (0.67 $\rightarrow$ 0.63), \textbf{Agreement Fraction} (0.68 $\rightarrow$ 0.66), \textbf{Switch Fraction} (0.34 $\rightarrow$ 0.28), \textbf{RAIR} (0.38 $\rightarrow$ 0.30), \textbf{RSR} (0.64 $\rightarrow$ 0.61).
This is evident in the condition \tutorialinc{}: \textbf{Accuracy} (0.68 $\rightarrow$ 0.63), \textbf{Agreement Fraction} (0.71 $\rightarrow$ 0.66), \textbf{Switch Fraction} (0.39 $\rightarrow$ 0.29), \textbf{RAIR} (0.43 $\rightarrow$ 0.29), \textbf{RSR} (0.59 $\rightarrow$ 0.61). When AI advice is in disagreement with users' initial decision, users tend to rely on themselves more than they should. This results in decreased (appropriate) reliance and accuracy.
In the deceptive review detection tasks, the AI system performs generally better than participants. The reduced reliance may help explain why we found a decreased accuracy on average.
% } \glcomment{To check whether we want to frame with under-reliance and underestimation of AI performance}
% ~\glcomment{Shall we still show detailed results of each debugging condition? Some condition showed insignificant change, but the trend is similar.}

% \begin{table}[htbp]
% 	\centering
% 	\caption{Wilcoxon signed ranks test results for \textbf{H2} on reliance-based dependent variables. ``$\dagger\dagger$'' indicates the effect of variable is significant at the level of 0.017 (adjusted alpha).}
% 	\label{tab:hypothesis-res-2}%
% 	\begin{footnotesize}
% 	\begin{tabular}{c | c  c c c }
% 	    \hline
% 	    \textbf{Dependent Variables}& $T$& $p$& $M \pm SD$(first)& $M \pm SD$(second)\\
% 	    \hline \hline
% 	    \textbf{Accuracy}& 3559& \textbf{.003}$^{\dagger\dagger}$& $0.67 \pm 0.16$& $0.62 \pm 0.15$\\
% 	    \rowcolor{gray!15}\textbf{Agreement Fraction}& 16677& \textbf{.001}$^{\dagger\dagger}$& $0.68 \pm 0.17$& $0.64 \pm 0.16$\\
% 	    \textbf{Switch Fraction}& 33403& \textbf{.001}$^{\dagger\dagger}$& $0.56 \pm 0.27$& $0.52 \pm 0.26$\\
% 	    \rowcolor{gray!15}\textbf{RAIR}& 60004& \textbf{.000}$^{\dagger\dagger}$& $0.51 \pm 0.30$& $0.46 \pm 0.29$\\
% 	    \textbf{RSR}& 82590& \textbf{.000}$^{\dagger\dagger}$& $0.54 \pm 0.34$& $0.49 \pm 0.34$\\
% 	    \hline
% 	\end{tabular}%
% 	\end{footnotesize}
% \end{table}

% \gladd{Further analysis about trust? shall we compare the trust change?}

\subsubsection{H3: ordering effect of debugging tasks}
For the analysis of the ordering effect, meanwhile mitigating the individual differences and learning effect brought by the eight tasks used in debugging phase, we compared the difference of reliance-based dependent variables (calculated with the difference between the second batch and the first batch) and user reliance on the second batch with participants of all conditions with Kruskal Wallis test. No significant difference is found with such comparisons. To compare the task working load brought by debugging intervention of different ordering, we conducted Kruskal-Wallis H-test on the six measures in the NASA-TLX questionnaire. No significant difference is found. Thus, \textbf{H3} is also not supported. 
% \glcomment{Add a table of each variable, at the last column show the order with significance (insignificant conditions are placed together)}

% \gladd{
To further look at how the ordering effect of debugging tasks affects the final performance of participants. We counted the participants who achieved an accuracy level above 80$\%$ (\ie compatible with or better than provided AI system) in the second task batch. After filtering out the participants who blindly rely on the AI system (\ie \textbf{Agreement Fraction} is 1.0), we found the number of participants in condition \tutorialdec{} (14) is clearly more than in condition \tutorialrandom{} (9) and \tutorialinc{} (9). In comparison, the number of participants who achieved an accuracy level above 80$\%$ in condition \control{} is 11. Although the ordering effect does not show a significant statistical difference, such an observation lends partial support to \textbf{H3}.%~\glcomment{Shall we only keep this, instead of the ``Thus, \textbf{H3} is also not supported. ''}

% \begin{table*}[htbp]
% 	\centering
% 	\caption{Kruskal-Wallis H-test results for \textbf{H3} on performance improvement of reliance-based dependent variables. ``$\dagger$'' indicates the effect of variable is significant at the level of 0.05.~\glcomment{Example of the conlusion of post-hoc analysis}}
% 	\label{tab:hypothesis-res-4-kruskal}%
% 	\begin{tabular}{c | c c | c}
% 	    \hline
% 	    \textbf{Dependent Variables}& $T$& $p$& Conclusion\\
% 	    \hline \hline
% 	    \textbf{Accuracy}& 4.15& \textbf{.042}$^\dagger$&\control < \tutorialrandom{}, \tutorialdec{} < \tutorialinc{}\\
% 	    \rowcolor{gray!15}\textbf{Agreement Fraction}& 0.98& .322&\\
% 	    \textbf{Switch Fraction}& 0.03& .854&\\
% 	    \rowcolor{gray!15}\textbf{RAIR}& 0.35& .555&\\
% 	    \textbf{RSR}& 1.16& .281& \\
% 	    \hline
% 	\end{tabular}%
% \end{table*}

% \begin{itemize}
%     \item Between-subjects study with Kruskal Wallis test. Compare the performance / assessment difference between first batch with the second batch.~\glcomment{In my view, such difference will not be greatly affected by individual difference. Because the difference try to assess the difference brought by tutorial.} Expectation: \control < \tutorialrandom{} < \tutorialdec{} < \tutorialinc{}. Although their assessment of AI performance and appropriate reliance get improved, the increasing impression order provides a better outcome. 
%     \item specific ordering may show relatively low cognitive load?
%     \item trust decrease?
% \end{itemize}

\subsection{Exploratory Analyses}
% \glcomment{How to analyze the confidence dynamics?}

\subsubsection{Trust Analysis} To explore whether our debugging intervention had any effect on user trust in AI system, we conducted Wilcoxon signed ranks test comparing the trust before and after the debugging intervention. 
% The results are: \textbf{TiA-R/C}, $T=4410$, $p=.046$; \textbf{TiA-U/P}, $T=4485$, $p=.932$; \textbf{TiA-IoD}, $T=1062$, $p=.017$; \textbf{TiA-Trust}, $T=2305$, $p=.045$. 
On average, there is a slight drop in assessed trust in automation subscales (\ie \textbf{TiA-R/C}, \textbf{TiA-U/P}, \textbf{TiA-IoD}, \textbf{TiA-Trust}) after the debugging intervention, but no statistically significant difference are found in test results. {This suggests that the designed debugging intervention can calibrate user reliance and estimation of AI performance without directly shaping their trust in the AI system.} 
% This suggests that the main impact of the debugging intervention was on the helping users calibrate their estimation of AI performance without directly shaping their trust in the AI system. 

\begin{table*}[hbpt]
% the environment \color{blue} change all cell color
	\centering
	\caption{{Kruskal-Wallis H-test results for user estimated trustworthiness and miscalibration of estimated performance based on performance quartiles. ``${\dagger\dagger}$'' indicates the effect of variable is significant at the level of 0.017. ``Top'', ``Middle'', and ``Bottom'' refers to participants in the top quartile, middle quartiles, and bottom quartile based on the performance of the first batch of tasks, respectively.}}
	\label{tab:qualtile_assessment}%
	\begin{small}
	\begin{tabular}{c | c c c c c| c}
	    \hline
	   % Hypothesis&	\multicolumn{4}{c|}{\textbf{H1}}& \multicolumn{4}{c}{\textbf{H2}} \\
	    % \textbf{Participants}&	\multicolumn{5}{c|}{\textbf{All}}& Post-hoc results\\
	    % \hline
	    \textbf{Dependent Variables}& $H$& $p$& $M \pm SD$(Top)& $M \pm SD$(Middle)& $M \pm SD$(Bottom)& Post-hoc results\\
	    \hline \hline
	\textbf{EAP}& 41.54& \textbf{<.001}$^{\dagger\dagger}$& $6.85 \pm 1.48$ & $5.84 \pm 1.91$&  $4.69 \pm 1.68$& Top > Middle > Bottom\\
	\textbf{ETP}& 15.85& \textbf{<.001}$^{\dagger\dagger}$& $7.29\pm 1.40$& $6.65 \pm 1.66$& $5.98\pm 1.96$& Top > Middle, Bottom\\
        \textbf{MAP}& 40.89& \textbf{<.001}$^{\dagger\dagger}$& $1.32\pm 1.33$& $2.29 \pm 1.74$& $3.31\pm 1.68$& Top < Middle < Bottom\\
        \textbf{MTP}& 22.67& \textbf{<.001}$^{\dagger\dagger}$& $1.46\pm 1.24$& $1.34 \pm 1.12$& $2.24\pm 1.28$& Top, Middle < Bottom\\
        \textbf{CCD}& 23.17& \textbf{<.001}$^{\dagger\dagger}$& $2.08\pm 1.70$& $2.06 \pm 1.78$& $0.86\pm 1.00$& Top, Middle > Bottom\\
	    \hline
	\end{tabular}%
	\end{small}
\end{table*}

\subsubsection{Covariates Impact on Trust and Reliance}\label{sec-covariate-impact}
To analyze the impact of covariates on user trust and reliance, we conducted the Spearman rank-order tests with covariates and the average trust and reliance-based dependent variables on two batches of tasks. 
The results show that, propensity to trust (\ie \textbf{TiA-PtT}) is the only factor which shows sinificant positive correlations with trust-based measures: \textbf{TiA-R/C} ($r(234)=0.270, p = .000$), \textbf{TiA-U/P} ($r(234)=0.165, p = .011$), \textbf{TiA-IoD} ($r(234)=0.234, p = .000$), \textbf{TiA-Trust} ($r(234)=0.303, p = .000$). 

\subsubsection{Users' Estimation of AI Trustworthiness}
{To further understand how users' estimation of AI trustworthiness affects their reliance and performance, we split the participants in all conditions into performance-based quartiles. 
To avoid the impact of debugging intervention, we only considered user performance in the first batch of tasks. 
The top quartile corresponds to those demonstrating high accuracy (top 25\%), the bottom quartile corresponds to those with low accuracy (bottom 25\%), and we combine the two quartiles in the middle comprising of participants with a medium level of performance in the first batch of tasks. 
To show how these participants differ in their appropriate reliance and estimation of AI trustworthiness, we adopted the Kruskal-Wallis H test to compare the estimated performance and their assessment of the AI system’s performance at the instance and global levels. 
% trustworthiness of the AI system (globally, Estimated AI performance; locally, CCD). 
Post-hoc Mann-Whitney tests using a Bonferroni-adjusted alpha level of 0.017 ( $\frac{0.05}{3}$) were used to make pairwise
comparisons of performance. 
Generally, participants in the high accuracy group showed more appropriate reliance (\ie \textbf{RAIR} and \textbf{RSR}) than the low accuracy group (with statistical significance). 
The results of user estimation of performance, AI trustworthiness, and miscalibration of performance are shown in Table~\ref{tab:qualtile_assessment}. 
Overall, participants in the high accuracy group showed significantly higher AI performance and team performance in comparison with the low accuracy group. Meanwhile, the high accuracy group also has a more precise estimation of AI performance and team performance (\ie significantly lower \textbf{MAP} and \textbf{MTP}) and makes more correct decisions confidently (significantly higher \textbf{CCD}). 
It also indicates that the underestimation of AI trustworthiness can be the main cause of the under-reliance, which results in lower accuracy.
}

\begin{figure}[h]
    \centering
    \includegraphics[scale=0.5]{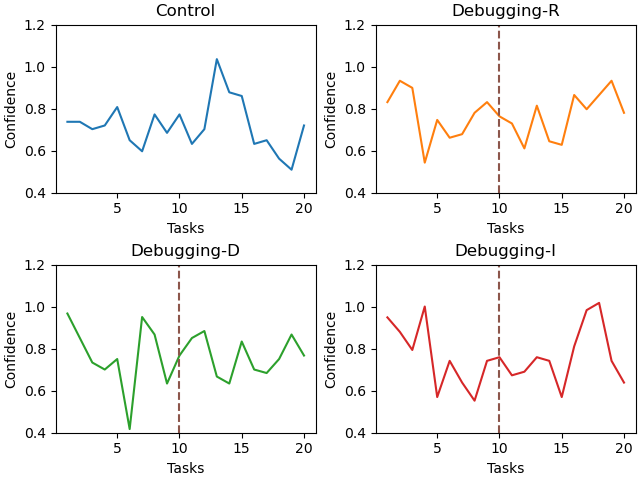}
    \caption{Illustration of dynamics of confidence change in the 20 tasks of each condition. The brown dashed line represents the debugging intervention.}
    \label{fig:confidence_dynamics}
\end{figure}

\begin{table}[htbp]
	\centering
	\caption{Reliance and confidence correlation.}
	\label{tab:reliance_pattern_confidence}%
% 	\begin{small}
	\begin{tabular}{c | c | c | c}
	    \hline
	    Pattern& \textbf{Dependent Variables}& $M$ & $SD$\\
	    \hline \hline
	    \multirow{5}{*}{Reliance}& \textbf{Initial agreement}& 0.38& 0.75\\
	    &\textbf{Initial disagreement}& -0.42& 0.99\\
	    &\textbf{Final agreement}& 0.23& 0.90\\
	    &\textbf{Final disagreement}& -0.44& 0.90\\
	    &\textbf{Switch behavior}& -0.32& 1.17\\
	    \hline
	     \multirow{4}{*}{\tabincell{c}{Appropriate\\ Reliance}}&\textbf{Positive AI reliance}& -0.34& 1.17\\
	    &\textbf{Negative AI reliance}& -0.23&1.18\\
	    &\textbf{Positive self-reliance}& -0.41&0.88\\
	    &\textbf{Negative self-reliance}& -0.48&0.89\\
	    \hline
	\end{tabular}%
% 	\end{small}
\end{table}

\subsubsection{Confidence Analysis}
We show the difference in confidence dynamics of four conditions in Figure~\ref{fig:confidence_dynamics}. On average, participants show positive confidence (above neutral) in their final decisions. After receiving the debugging intervention, both \tutorialinc{} and \tutorialrandom{} conditions showed decreased confidence, but it comes back to the average level soon and keeps vibrating around it. By contrast, participants in condition \tutorialdec{} showed increased confidence after the debugging intervention and keeps relatively stable compared with all other conditions.

% \begin{figure}[h]
%     \centering
%     \includegraphics[scale=0.5]{figures/confidence_dynamics_line.png}
%     \caption{Illustration of dynamics of confidence change in the 20 tasks of each condition.}
%     \label{fig:confidence_dynamics_2}
% \end{figure}

% Confidence dynamics with initial agreement, and initial disagreement. \gladd{With 101 valid users, we calculated the change of confidence under different reliance patterns. We found that participants who agree with AI systems initially, will show increased confidence after receiving the advice and explanation. By contrast, their confidence decrease when disagree with AI advice. For the four patterns considered in appropriate reliance measures, only negative AI reliance witnessed a slight confidence increase.}

We calculated the confidence change after receiving AI advice based on nine different reliance patterns: whether initial decision agrees with AI advice, whether final decision agrees with AI advice, switch behavior, and four reliance patterns considered in calculating appropriate reliance (see Table~\ref{tab:reliance_patterns}). 
The results are shown in Table~\ref{tab:reliance_pattern_confidence}. 
In general, participants indicated increased confidence when AI advice agreed with their initial decision, and showed decreased confidence when AI advice disagreed with their initial decision. And even if participants choose to switch to AI advice given initial disagreement, they tend to show decreased confidence in the final decision. Considering the four patterns in calculating appropriate reliance, users' confidence drop seems to be more severe when insisting on their own decision, compared with adopting AI advice.
% ~\glcomment{Shall we mention, we compared them pairwise, and obtained one confidence ordering of these reliance patterns? Or it is fine that we just use the M value to indicate the different trend}

% \begin{table}[h!]
% 	\centering
% 	\caption{Reliance and confidence correlation.}
% 	\label{tab:reliance_pattern_confidence}%
% % 	\begin{small}
% 	\begin{tabular}{c | c | c | c}
% 	    \hline
% 	    Pattern& \textbf{Dependent Variables}& $M$ & $SD$\\
% 	    \hline \hline
% 	    \multirow{5}{*}{Reliance}& \textbf{Initial agreement}& 0.38& 0.75\\
% 	    &\textbf{Initial disagreement}& -0.42& 0.99\\
% 	    &\textbf{Final agreement}& 0.23& 0.90\\
% 	    &\textbf{Final disagreement}& -0.44& 0.90\\
% 	    &\textbf{Switch behavior}& -0.32& 1.17\\
% 	    \hline
% 	     \multirow{4}{*}{\tabincell{c}{Appropriate\\ Reliance}}&\textbf{Positive AI reliance}& -0.34& 1.17\\
% 	    &\textbf{Negative AI reliance}& -0.23&1.18\\
% 	    &\textbf{Positive self-reliance}& -0.41&0.88\\
% 	    &\textbf{Negative self-reliance}& -0.48&0.89\\
% 	    \hline
% 	\end{tabular}%
% % 	\end{small}
% \end{table}

% Confidence dynamics associated with appropriate or inappropriate reliance patterns.

%% file: sections/sec-discussion.tex
\section{Discussion}
% \glcomment{How to add discussion about plausibility}
\subsection{Key Findings}
% With the wish of teaching participants how to appropriately rely on AI systems, 
In order to promote appropriate reliance on AI systems by calibrating user estimation of AI performance, we proposed a debugging intervention to educate participants that AI systems are not always reliable and that the explanations may also not always be informative. 
{We hypothesized that the proposed debugging intervention could improve critical thinking about the AI system, which can facilitate appropriate reliance on the AI system.} 
As opposed to our hypotheses, such a debugging intervention fails to calibrate participants' estimation of AI performance at both the global and local levels. Participants tended to rely less on the AI system after receiving the debugging intervention. 
{Through an exploratory analysis based on different performance quartiles, we found that participants who performed worse in our study tended to underestimate AI performance. 
Thus, they achieved suboptimal team performance, which is largely impacted by the under-reliance on the AI system. 
These findings can also be explained using the lens of plausibility of the XAI intervention. 
According to Jin \etal~\cite{jin2023rethinking}, plausibility  can substantially affect user perceived trustworthiness of the AI system. 
The debugging intervention may make the XAI (\ie text highlights in our study) less plausible to users, which results in more tendency to underestimate AI performance.}

% While no significant difference was found between the different ordering of debugging tasks across experimental conditions, participants who were exposed to the weakness of the AI system at the beginning of the debugging intervention, showed a more obvious tendency to disuse the AI system, and such under-reliance was found to result in sub-optimal team performance. 
{In our study, no significant difference was found between the different ordering of debugging tasks across experimental conditions. However, participants who were exposed to the weakness of the AI system at the beginning of the debugging intervention, showed a more obvious tendency to disuse the AI system. Such under-reliance was found to result in sub-optimal team performance.}
% No significant difference is found due to ordering effect of debugging tasks which are used to show both strength and weakness of AI systems. 
% between debugging intervention with different ordering of tasks to show both strength and weakness of AI systems, 
% However, we did find that when presenting the participants with the weakness of AI system in the beginning of debugging intervention, participants tend to rely less on AI system in the later collaboration with the same AI system.
% As a result, they tend to rely less on AI systems, especially in condition where the weakness of AI system is exposed to participants in the beginning. 
This finding is in line with recent work that has uncovered similar %existing finding of cognitive biases due to 
ordering effects and cognitive biases influencing outcomes in human interaction with intelligent systems~\cite{Tolmeijer-UMAP-2021,nourani2021anchoring}: a bad first impression %brought by the revealed weakness of 
of an AI system can lead to an underestimation of AI competence and reduced reliance on the system.

\paratitle{Confidence Analysis}. We calculated the confidence change after receiving AI advice based on nine different reliance patterns: whether initial decision agrees with AI advice, whether final decision agrees with AI advice, switch behavior, and four reliance patterns considered in calculating appropriate reliance (see appendix). 
% The results are shown in Table~\ref{tab:reliance_pattern_confidence}. 
In general, participants indicated increased confidence when AI advice agreed with their initial decision ($+0.38$ on average), and showed decreased confidence when AI advice disagreed with their initial decision ($-0.42$ on average). And even if participants choose to switch to AI advice given initial disagreement, they tend to show decreased confidence in the final decision ($-0.32$ on average). Considering the four patterns in calculating appropriate reliance, users' confidence drop seems to be more severe when insisting on their own decision, compared with adopting AI advice.

In further analysis of covariates {(cf. Sec~\ref{sec-covariate-impact})}, we found that general propensity to trust shows a positive correlation with all trust subscales. However, no significant correlations were found between the propensity to trust and reliance, which indicates that the increased trust due to the propensity to trust does not translate to reliance behaviors. 
Meanwhile, the confidence dynamics in different reliance patterns showed that AI advice may amplify the confidence of user decisions when in agreement and decrease user confidence when in disagreement. 
Under disagreement, users appear to rely more on themselves (\ie indicated by confidence decrease), as opposed to adopting AI advice.

\subsection{Implications}
% \begin{itemize}
%     \item When designing user tutorials about AI strength and weakness, we should avoid leave a bad first impression.
%     \item Participants' overestimation of themselves and underestimation of AI performance 
% \end{itemize}
Our findings suggest that the debugging intervention and similar interventions with training purposes (\eg user tutorial) may suffer from the cognitive bias brought by the ordering effect within such interventions. If we want to use such interventions to show users both strength and weakness of AI systems, we should avoid leaving users with a bad first impression of the weakness of the AI system. Meanwhile, in our study, participants tend to be optimistic about the team performance while underestimating the AI performance. It is possibly caused by meta cognitive bias --- Dunning-Kruger effect~\cite{kruger1999unskilled,he2023knowing}. 
{According to previous work~\cite{kruger1999unskilled}, Dunning-Kruger Effect is mainly triggered among less-competent individuals overestimating their own competence/performance in a task. 
In our study, we found that less-competent individuals showed a greater tendency to underestimate the AI performance and make fewer correct decisions with confidence (see Table~\ref{tab:estimation}). This indicates that the underestimation of AI systems can also contribute to under-reliance in the context of human-AI decision making. According to He \etal~\cite{he2023knowing}, an overestimation of self-competence can result in under-reliance on the AI system. 
Both the overestimation of self-competence and the underestimation of AI competence can contribute to an illusion of superior competence over the AI system. 
As a result, users with such an illusion tend to disuse the AI system. 
To conclude whether the underestimation of AI performance plays a role in triggering the Dunning-Kruger effect in the context of human-AI decision making, more work is required in the future. 
% Future work 
}

% For the wish to promote appropriate reliance, we need to help participants understand AI systems from the standpoints of both their strengths and their weaknesses. 
% It is also highly important to help users better understand their own strengths and weakness with respect to the task at hand. 
% {With such a comprehensive understanding, users have a larger chance to calibrate their estimation of AI trustworthiness and their competence. This can form a vital basis for the appropriate use of AI advice.} 
% Our findings can inform the future design of training interventions to promote appropriate reliance.

Through our study, we also found that the reliance patterns (\eg agreement, disagreement) have a clear correlation with user confidence change. 
When the AI system disagrees with human initial decision, decision makers' confidence shows a clear decrease. 
And compared with insisting on their own decision, they may have higher confidence when giving agency to AI advice. 
Such observation may be a dangerous signal for appropriate reliance. 
Further research is required to explore how to keep user confidence on themselves when exposed to a disagreement from an AI system.

\subsection{Caveats and Limitations}
{Our debugging intervention may have left participants with a negative impression of the AI system, which could irreversibly harm the trust and reliance on the system (as shown by prior literature exploring first impressions of AI systems~\cite{Tolmeijer-UMAP-2021}). 
To make the debugging intervention more effective in building up critical mindsets and facilitating appropriate reliance, future research can explore how to avoid such side-effects. The high difficulty of the task and the debugging intervention may have influenced our findings. In a highly complex task, crowd workers may not be patient and engaged enough to fully understand the AI system at both the global and local level. Although we only focus on one specific task to verify the effectiveness of the proposed debugging intervention, such an application-grounded evaluation is still highly valuable~\cite{doshi2017towards}.
In this work, we used a rigorous setup to explore the effectiveness of a debugging intervention, which can inform the future design of effective interventions for better human-AI collaboration.}
% \glcomment{We will add a limitations subsection and missing reference. We agree that the high difficulty of the task and debugging intervention may have influenced our findings; yet this is a stringent setup to explore the benefit of the intervention. (2) We agree that manipulating user impressions is unethical. We meant that in a debugging intervention, creating a highly negative impression of an AI system could irreversibly harm the trust and reliance on the system. We will clarify this in the revision.}

\paratitle{Potential Bias}. 
As pointed out by Draws \etal~\cite{draws2021checklist}, cognitive biases introduced by task design and workflow may have negative impact on crowdsourcing experiments. 
With the help of Cognitive Biases Checklist introduced~\cite{draws2021checklist}, we analyzed potential bias in our study. 
\textit{Self-interest bias} is possible, because crowd workers we recruited from the Prolific platform are motivated the monetary compensation. Thus, it would be challenging to keep participants engaged in the debugging intervention and highly motivated to learn from the weakness of AI system. That could be also potential reason why the debugging intervention does not work as expected. To alleviate any participants with low effort results, we put attention checks to remove ineligible participants from our study. The observation of reduced reliance brought by bad first impression also happens with \textit{Anchoring Effect}. Meanwhile, the participants generally under-estimate the AI performance and believe they can outperform AI system, which also may fall into \textit{Overconfidence or Optimism Bias}.

%% file: sections/sec-con.tex
\section{conclusion}
In this paper, we present an empirical study to understand the impact of the debugging intervention on the estimation of AI performance and user reliance on the AI system. 
Our results suggest that we should be careful in presenting the weakness of the AI system to users, to avoid any anchoring effect which may result in under-reliance. 
While our experimental results do not provide support to our original hypotheses, we can not fully reach a conclusion that debugging intervention does not help with facilitating appropriate reliance on the AI system. 
Future work may explore how to mitigate potential bias brought by the users' overestimation of themselves along with the underestimation of AI performance. 
Meanwhile, our observations of confidence dynamics in different reliance patterns also provide insights for future study of human-AI decision making.

%% file: sections/appendix.tex
\appendix

\section{Appendix}
\subsection{Experimental Details}
\label{sec-appendix-implementation}
\paratitle{Guidelines}. Following Lai \etal~\cite{Lai-CHI-2020}, we provided the following guidelines in the user study:

\begin{itemize}
    \item Deceptive reviews tend to focus on aspects that are external to the hotel being reviewed, \eg husband, business, vacation.
    \item Deceptive reviews tend to contain more emotional terms; positive deceptive reviews are generally more positive and negative deceptive reviews are more negative than genuine reviews.
    \item Genuine reviews tend to include more sensorial and concrete language, in particular, genuine reviews are more specific about spatial configurations, \eg small, bathroom, on, location.
    \item Deceptive reviews tend to contain more verbs, \eg eat, sleep, stay.
    \item Deceptive reviews tend to contain more superlatives, \eg cleanest, worst, best.
    \item Deceptive reviews tend to contain more pre-determiners, which are normally placed before an indefinite article + adjective + noun, \eg what a lovely day!
\end{itemize}

\paratitle{Timer}. Besides attention checks, we also add a timer to ensure each participant spends enough time on the questionnaire, task instruction, and decision tasks.  A conservative estimate through trial runs reflected that participants would take at least 25 seconds to complete each decision task and 30 seconds to complete each debugging task. We reduced the time for the decision making in the second stage to 15 seconds. 
% Since attention check pages do not require deliberation, we reduced that time to 5 seconds.

\paratitle{Qualification Test}. To ensure participants carefully read the task instruction and understand the task, we used two questions for the qualification test. 
\begin{itemize}
    \item In this study, the deceptive reviews written by? Option 1: An AI system, option 2: People without actual experience.
    \item Indicate whether the following statement is true or false: "Guidelines are provided for finding deceptive reviews". Option 1: True, option 2: False.
\end{itemize}

\begin{figure}[h]
    \centering
    \includegraphics[width=0.48\textwidth]{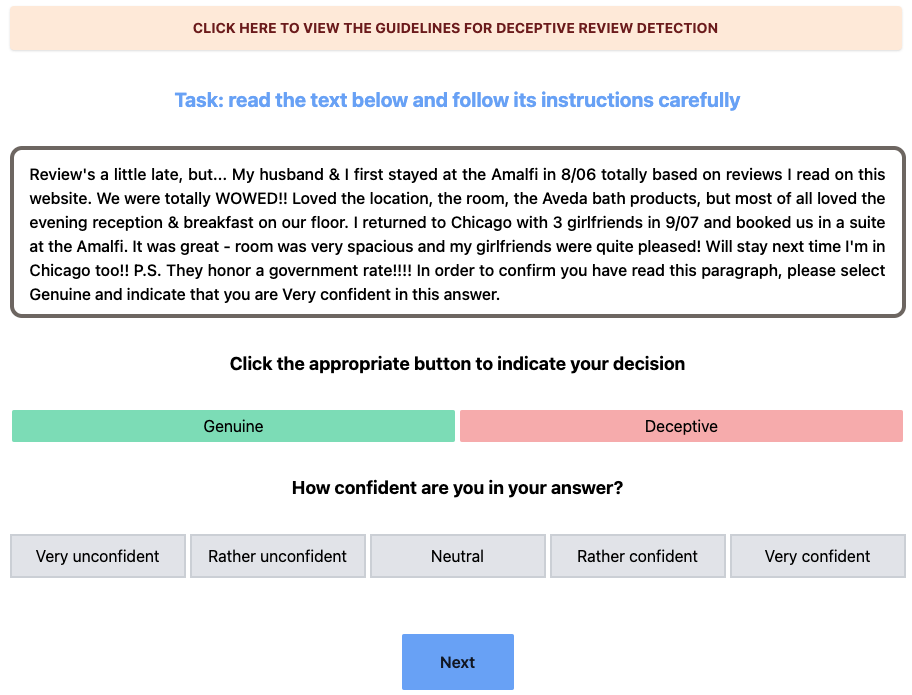}
    \caption{Screenshot of attention check in decision tasks.}
    \label{fig:attention_check}
\end{figure}

\paratitle{Attention Checks}. To prevent participants from providing low-effort results in questionnaires and decision tasks, we add attention check tasks that are similar to normal ones. For example, we asked participants to select a specified option in the questionnaire. One example of attention check in decision tasks is shown in Figure~\ref{fig:attention_check}. To ensure participants read the hotel review with attention, we put the instruction in the last sentence to select a specific option (\eg ``In order to confirm you have read this paragraph, please select Genuine and indicate that you are Very confident in this answer.''). We have such attention checks in the middle of each task batch, long questionnaires, and the debugging intervention.

